\pdfoutput=1
\documentclass{article}
\usepackage{le_preprint,times}

\usepackage{amsmath,amsfonts,bm}

\def\eqref#1{equation~\ref{#1}}

\def\1{\bm{1}}

\DeclareMathAlphabet{\mathsfit}{\encodingdefault}{\sfdefault}{m}{sl}
\SetMathAlphabet{\mathsfit}{bold}{\encodingdefault}{\sfdefault}{bx}{n}

\newcommand{\E}{\mathbb{E}}

\usepackage{hyperref}
\usepackage{url}

\title{Logical Equivariance in Diffusion Language Models}

\author{Sunwoo Hong$^{1,2*\ddagger}$\quad
Seo Hyun Kim$^{1,2*\ddagger}$\quad
Chen-Hao Chao$^{2}$\quad
Younwoo Choi$^{2}$\\
\bfseries Se-Young Yun$^{1\dagger}$\quad
Rahul G. Krishnan$^{2\dagger}$\\[2pt]
\normalfont$^{1}$KAIST AI\quad $^{2}$University of Toronto \& Vector Institute\\
\normalfont\texttt{\char`\{sunwoo, shkimsally\char`\}@kaist.ac.kr}}

\usepackage{microtype}
\usepackage{graphicx}
\usepackage{subcaption}
\usepackage{booktabs}

\usepackage{hyperref}

\usepackage{amsmath, amssymb, amsthm, mathtools}
\usepackage{bm}
\usepackage{multirow}
\usepackage{enumitem}
\usepackage[most]{tcolorbox}
\usepackage{makecell}
\usepackage{tabularx}
\usepackage{array}
\usepackage{float}
\usepackage[flushleft]{threeparttable}
\usepackage[ruled,vlined,linesnumbered]{algorithm2e}
\usepackage{placeins}
\usepackage{listings}
\usepackage{inconsolata}
\usepackage{xspace}
\usepackage{newunicodechar}
\usepackage{pifont}
\usepackage[capitalize,noabbrev]{cleveref}

\usepackage[table]{xcolor}
\definecolor{myteal}{HTML}{069494}

\newunicodechar{ }{\,}
\SetKwInput{Input}{Input}
\SetKwInput{Output}{Output}
\SetKw{Initialize}{Initialize}
\DontPrintSemicolon
\SetKwBlock{Block}{\{}{\}}

\theoremstyle{plain}
\newtheorem{theorem}{Theorem}[section]
\newtheorem{proposition}[theorem]{Proposition}
\newtheorem{lemma}[theorem]{Lemma}
\newtheorem{corollary}[theorem]{Corollary}
\theoremstyle{definition}

\theoremstyle{remark}

\hypersetup{pdfauthor={Sunwoo Hong, Seo Hyun Kim, Chen-Hao Chao, Younwoo Choi, Se-Young Yun, Rahul G. Krishnan},pdftitle={Logical Equivariance in Diffusion Language Models}}
\makeatletter
\newcommand{\authornote}[1]{\begingroup\insert\footins{\footnotesize
  \interlinepenalty\interfootnotelinepenalty
  \splittopskip\footnotesep\splitmaxdepth\dp\strutbox
  \floatingpenalty\@MM\hsize\columnwidth\@parboxrestore
  \color@begingroup\noindent\rule\z@\footnotesep\ignorespaces#1\@finalstrut\strutbox
  \color@endgroup}\endgroup}
\makeatother
\begin{document}
\maketitle
\authornote{$^*$Equal contribution.\quad $^\dagger$Corresponding authors.\quad $^\ddagger$Work done as visiting researchers at the University of Toronto and the Vector Institute.}
\begin{abstract}
Rewriting a problem without changing its answer can change whether a language model solves it, even when average accuracy barely moves. We study this as \textbf{logical equivariance} through exact transported targets and paired evaluations on Sudoku, code, and mathematics. Same-input decoding controls distinguish presentation-dependent correctness from repeated-decoding variability in four diffusion language models. Autoregressive baselines exhibit the phenomenon as well. We prove that exact grid equivariance restricts row-major relative-score attention to two or three offset classes. In post-training, Reasoning-State Masking (RSM) selects conditional inputs and presentation randomization (PR) varies their equivalent views. Matching puzzle frequency, mask density, and per-target supervision preserves mean-success interactions of 1.51 and 2.45 percentage points at 50M and 100M parameters. The effect also survives an unbiased success-dispersion estimator and a token-marginal sampling calibration. We characterize a state-dependent presentation-reuse term in gradient covariance and test it by changing the refresh schedule. The combined recipe improves mean success by 2.10 points on matched mazes and raises all-view correctness at 8B from 47.7\% to 86.1\%. Interaction, endpoint utility, and recipe-level transfer are evaluated separately. Our project page is available at {\hypersetup{pdfborder={0 0 0}}\href{https://sunwoohong.github.io/logical-equivariance/}{\textcolor{myteal}{\nolinkurl{sunwoohong.github.io/logical-equivariance}}}}.
\end{abstract}
\section{Introduction}
\label{sec:introduction}

\begin{figure}[t]
\centering
\includegraphics[width=0.97\textwidth]{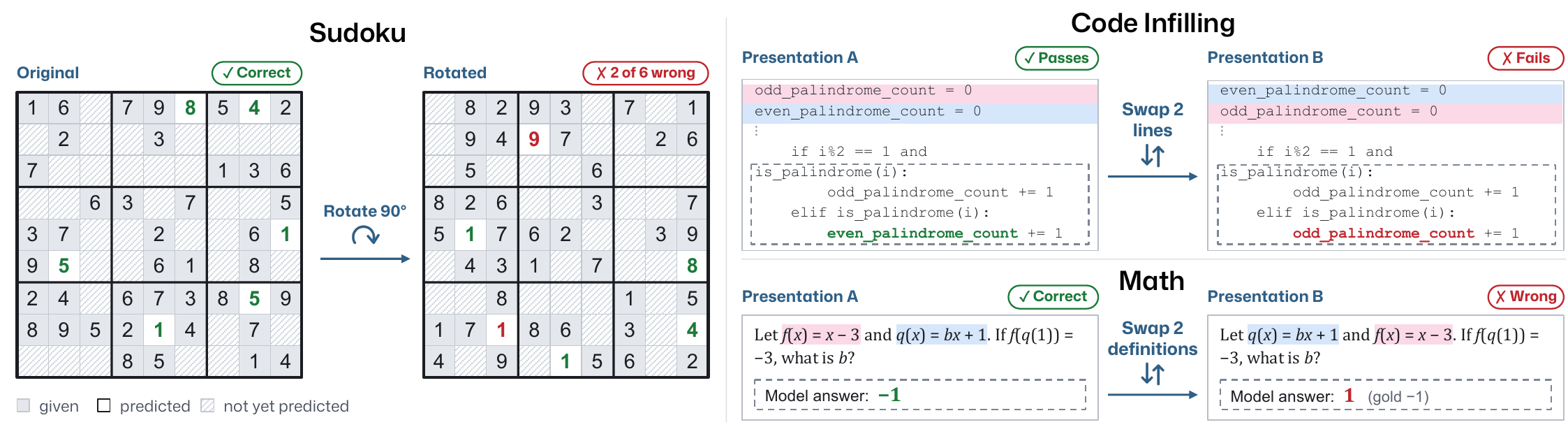}
\caption{\textbf{A correct answer does not survive an equivalent change of presentation.}
Rotating a Sudoku board, swapping two independent lines of code, or reordering two definitions in a math problem leaves the task and its answer unchanged, yet LLaDA-8B solves one presentation and fails the other.
Accuracy on a single presentation cannot reveal this: whether a problem counts as solved depends on how it happens to be written.}
\label{fig:motivation}
\end{figure}

The same problem can be written in many ways. Variables can take different names, and statements or conditions can appear in a different order, without changing what is being asked. Current models do not always meet this expectation~\citep{lu2022fantasticallyorderedpromptsthem,liu2023lostmiddlelanguagemodels}. In Figure~\ref{fig:motivation}, LLaDA-8B~\citep{nie2025large} answers a Sudoku puzzle, a code completion, and a math question correctly in one form but fails after an equivalent rewrite. The form of a correct answer may change with the input, but its correctness should not. Behavioral testing and code-generation robustness studies have highlighted the importance of this requirement~\citep{ribeiro2020accuracybehavioraltestingnlp,wang2022recoderobustnessevaluationcode}.

Average accuracy can hide how often these successes are lost. In our MATH-500 evaluation, consistent variable renaming raises LLaDA-8B's accuracy from 28.5\% to 30.2\%, even though correctness changes on 17.9\% of instances. Renaming breaks 8.1\% of all instances, more than a quarter of those originally solved, while repairing another 9.8\%. These changes largely cancel in the average, but not for the user whose problem is no longer solved.

Preserving success across equivalent inputs is a question of generalization: can a learned solving ability transfer when notation changes while the defining relations remain fixed? This also provides a practical reason to seek internal representations that capture these relations~\citep{cohen2016groupequivariantconvolutionalnetworks,bronstein2021geometric}. We study this observable property as \emph{logical equivariance} (LE), where predictions transform consistently with specified answer-preserving input changes. Tracking each problem across its views distinguishes favorable-view success from consistent correctness in every presentation, building on consistency evaluation~\citep{elazar2021measuringimprovingconsistencypretrained,Cho_2025}.

Diffusion language models make this learning question concrete~\citep{austin2021structured,ou2024your,sahoo2024simple,shi2024simplified,nie2025large,ye2025dream7bdiffusionlarge}. Their masked-denoising objective trains on partially observed inputs, so the masking distribution determines which conditional tasks the model practices. Flexible decoding can choose which position to fill next~\citep{kim2025train,liu2024think,peng2025path,kim2026klass,ni2026flexibility}; presentation changes instead alter which problem object occupies each position. We investigate how the selection of partial inputs and exposure to equivalent views affect reliable solving.

Building on Sudoku reasoning benchmarks~\citep{seely2025sudokubenchevaluatingcreativereasoning}, our \textsc{Sudoku-AC-Orbit} benchmark supplies an exact test: its targets are logically forced and transport under all eight rotations and reflections. Together with paired code-infilling and mathematics evaluations, it reveals presentation-dependent correctness in four large dLLMs. On Sudoku-AC32, Dream-7B, iLLaDA-8B, and LLaDA-MoE each solve about 99\% of states in at least one view, but only 12.9--23.8\% in every view.

We also characterize the positional information compatible with this symmetry. A row-major relative-score attention head~\citep{vaswani2017attention} is exactly $D_4$-equivariant under a position-only action only if its offset-indexed score functions collapse to two classes on even-sized grids and three on odd-sized grids (Theorem~\ref{thm:rigidity}). Standard row-major RoPE does not impose these ties~\citep{su2024roformer}; we therefore investigate presentation stability through post-training while keeping the backbone unchanged.

We combine \emph{Reasoning-State Masking} (RSM), which supplies partial inputs encountered during a reasoning process and densely supervises unresolved targets, with \emph{presentation randomization} (PR). The question is not whether augmentation can help, but whether its effect changes with the conditional tasks practiced. At 100M, with puzzle frequency, hidden-cell count, and target supervision matched, PR changes mean success by $-1.00$ points on random states and $+1.44$ on reasoning states.

Our contributions are:
\begin{itemize}
\item \textbf{Logical equivariance.} We formalize consistency under answer-preserving rewrites and build forced-target orbits on which targets transport exactly. Same-input controls and an unbiased success-dispersion estimator separate presentation dependence from decoding variability in four dLLMs, and AR baselines show the same phenomenon.
\item \textbf{Serialization rigidity.} We prove that exact grid symmetry restricts row-major relative-score attention, including RoPE, to two or three offset classes, so the symmetry must be learned.
\item \textbf{State-dependent augmentation.} Presentation randomization is not uniformly beneficial: under matched puzzle frequency, mask density, and per-target supervision, it lowers mean success on random states and raises it on reasoning states. A presentation-reuse term in gradient covariance, larger for reasoning states, explains this, and refresh-schedule interventions support it. Designed jointly, RSM+PR improves solving over matched random-state training on Sudoku and mazes.
\end{itemize}

\section{Related Work}
\label{sec:related}
\paragraph{Consistency and structured evaluation.}
\citet{ribeiro2020accuracybehavioraltestingnlp} introduced behavioral testing as an alternative to aggregate accuracy metrics, while \citet{elazar2021measuringimprovingconsistencypretrained} evaluated consistency across meaning-preserving inputs, tracking both single-instance success and all-instance correctness. Similarly, \citet{sclar2024quantifyinglanguagemodelssensitivity} analyzed prompt-format sensitivity, and \citet{wang2022recoderobustnessevaluationcode} assessed code generation robustness under semantics-preserving edits.
\citet{lai2026multidimensional} study consistency across prompt order, phrasing, and language. Our AR comparison is compatible with this established sensitivity. It is not a claim that the failure is diffusion-exclusive.

We extend this evaluation perspective to diffusion reasoning models under exact target transportation. While Sudoku variants modify core task constraints~\citep{seely2025sudokubenchevaluatingcreativereasoning}, our within-orbit comparisons hold them strictly invariant.

\paragraph{Selecting conditional training tasks.}
Masked diffusion learns conditional token predictors~\citep{sahoo2024simple,nie2025large, ye2025dream7bdiffusionlarge}. \citet{kim2025train} analyze masking-induced prediction tasks and adaptive decoding, while \citet{shih2022traininginferenceanyorderautoregressive} select and reweight conditionals for any-order autoregressive models. \citet{shah2024causallanguagemodelingelicit} supervise solver-ordered output sequences; AGDO uses attention-derived dependencies for denoising and post-training~\citep{deng2026fullyrandommaskingattentionguided}. RSM instead uses a reasoning process to select partial input states and retains dense completion of unresolved targets. We study how this state distribution interacts with equivalent-view exposure.

\paragraph{Equivariance and augmentation.}
Parameter sharing and compatible positional encodings can enforce equivariance~\citep{ravanbakhsh2017equivarianceparametersharing,romero2021groupequivariantstandaloneselfattention}. Our theorem identifies the additional restrictions imposed by row-major offset sharing. The supplementary loss decomposition follows the logarithmic-opinion-pool identity of \citet{3008904.3008942}.
Our finite-sample measurement uses an independent-product U-statistic, a standard distribution-comparison construction~\citep{gretton2012kernel}. The application here separates success-probability dispersion, token marginals, and full-output equivariance. The learning-signal analysis complements the group-averaged augmentation perspective~\citep{chen2020group} by making the state sampler and presentation-reuse schedule explicit.

\section{Logical Equivariance}
\label{sec:le}
\subsection{Equivalent problems and transported outputs}
Let a finite group $G$ act bijectively on inputs $\mathcal X$ and outputs $\mathcal Y$, preserving the task's correctness relation. Include an invalid-output symbol $\bot$ with $g\bot=\bot$. A deterministic decoder $D$ is logically equivariant when $D(gx)=gD(x)$. This property is distinct from correctness: an equivariant model may be consistently wrong. For a stochastic decoder with law $K_\theta(\cdot\mid x)$, define
\begin{equation}
P_g(\cdot\mid x)=g^{-1}_{\#}K_\theta(\cdot\mid gx),
\qquad
\text{LE on }x\iff P_g=P_h\ \text{for every }g,h\in G,
\label{eq:le}
\end{equation}
where $g_{\#}$ denotes pushforward. Canonicalization removes the prescribed transformation of the answer. We evaluate one base problem together with its views, preserving the pairing between presentations. For code and mathematics transformations that are not a closed group, we use validated original--transformed pairs and the corresponding correctness evaluator.

\subsection{An exact orbit with forced targets}
An anchored-completion state $x_{\mathrm{AC}}(A,U,R)$ specifies visible value-assigned cells $A$, scored cells $U$, and other unresolved cells $R$. If $\mathcal S(A)$ is the set of valid Sudoku completions consistent with $A$, we retain states satisfying
\begin{equation}
\{s_U:s\in\mathcal S(A)\}=\{s_U^\star\}.
\label{eq:forced}
\end{equation}
Thus all valid completions agree on the scored answer. Rotations and reflections preserve rows, columns, and boxes, so $\mathcal S(gA)=g\mathcal S(A)$ and the unique target assignment transports with $g$. Appendix~\ref{app:forced} proves this claim. We generate target sets from contiguous solver-trace windows: \textsc{AC}$k$ denotes $k$ scored cells, with earlier assignments exposed as anchors.

\subsection{Correctness across presentations}
\label{subsec:metrics}
For the recorded prediction in view $g$, let $a_g(x)$ indicate correctness after canonicalization. With $m=|G|$, define
\begin{equation}
\mathrm{Acc}=\E_x\frac1m\sum_g a_g(x),\quad
\mathrm{ANY}=\E_x\max_g a_g(x),\quad
\mathrm{ALL}=\E_x\min_g a_g(x).
\label{eq:anyall}
\end{equation}
For a fixed deterministic decoder, ANY and ALL measure favorable-view and all-view correctness. For two views, let $b$ and $r$ be the fractions of all instances broken and repaired by the transformation. Then
\begin{equation}
\mathrm{Acc}_{g}-\mathrm{Acc}_{e}=r-b,
\qquad
\mathrm{ANY}-\mathrm{ALL}=r+b.
\label{eq:two-view-identity}
\end{equation}
Net accuracy change can therefore be small even when many instances change correctness. With sampled predictions, these formulas describe the recorded outcomes rather than equality of the output laws.

For stochastic decoding, let $p_g(x)$ denote task-success probability. We estimate it from $K$ samples per view and report
\begin{equation}
\begin{aligned}
\mathrm{MeanSucc}&=\E_x\frac1m\sum_g p_g(x),& C_{\max}&=\E_x\max_g p_g(x),\\
R_{\min}&=\E_x\min_g p_g(x),&\mathrm{sECR}&=R_{\min}/C_{\max}.
\end{aligned}
\label{eq:stochastic}
\end{equation}
Hats denote finite-$K$ estimates, and the ratio is defined when $C_{\max}>0$. Appendix Figure~\ref{fig:stochastic_metrics} illustrates how these quantities are computed from per-view success rates. Extrema over sampled success rates are biased, so the Sudoku recipe study includes a canonical pseudo-orbit control (Appendix~\ref{app:finite-k}). Correctness summaries are complemented by token-marginal view leakage $\widehat L_{\mathrm{tok}}$. Its population counterpart averages KL divergence to the across-view mixture, and full-output leakage is zero exactly when Eq.~\ref{eq:le} holds (Appendix~\ref{app:view-spectrum}). Table~\ref{tab:orbit-metrics-summary} lists every metric used in the paper with its definition.
\paragraph{A population quantity without sampled extrema.}
For a fixed problem let $p=(p_1,\ldots,p_m)^\top$, $\bar p=m^{-1}\sum_g p_g$, and $H=I_m-\mathbf1\mathbf1^\top/m$. Define
\begin{equation}
 V(x)=\frac1m\sum_g(p_g(x)-\bar p(x))^2,\qquad
 \widehat V(x)=\frac{\sum_{k\ne\ell}S^{(k)\top}HS^{(\ell)}}{mK(K-1)}.
 \label{eq:unbiased-v}
\end{equation}
If the success-indicator vectors $S^{(k)}$ are independent across replicates with common mean $p$, then $\E[\widehat V\mid x]=V(x)$ for every $K\ge2$. Within-replicate view dependence is allowed. We retain negative estimates. Appendix~\ref{app:unbiased-proof} proves the identity and distinguishes its assumptions from the stronger conditional-null model. For fixed trained arms, the favorable interaction is $J=\E_x[(V_C-V_D)-(V_A-V_B)]$. Positive $J$ means a greater PR-induced dispersion reduction under RSM, not superiority of the augmented endpoint or equality of full output laws.

\section{Method}
\label{sec:method}
RSM changes which conditional tasks receive supervision. PR changes how those tasks are presented. We retain the general state-selection objective and dense completion of unresolved targets. The structural result characterizes what the unchanged positional component can guarantee.

\subsection{Reasoning-State Masking}
\label{sec:lts}
Let $e=(c,y^\star)$ contain a problem specification $c$ and a structured reference $y^\star$ indexed by $\mathcal I_e$. A task-specific state generator $\mathcal O$ supplies a valid reasoning trajectory $(s_0,\ldots,s_T)$. Each nonterminal state defines an input $x_t=\operatorname{Enc}(c,s_t)$ and unresolved target indices $H_t\subseteq\mathcal I_e$. With example sampler $\nu$ and state sampler $\pi_e$, RSM optimizes
\begin{equation}
\mathcal L_{\mathrm{RSM}}(\theta)
=\E_{e\sim\nu,\,t\sim\pi_e}
\frac1{|H_t|}\sum_{i\in H_t}-\log p_{\theta,i}(y_i^\star\mid x_t).
\label{eq:lts_general}
\end{equation}
The reasoning process selects the next \emph{input state}. Every unresolved target remains supervised. For multi-token fields, the encoding and token-level normalization are part of the task specification.

\paragraph{Which conditional tasks are practiced?}
RSM changes the distribution of visible information, not the completion target or inference budget. Random partial observations of a correct reference may themselves be logically consistent. RSM selects states encountered by a specified reasoning process. Sound deductions can make consequences explicit without changing the feasible answers. Appendix~\ref{app:general_lts} formalizes this distinction. The relevant learning question is which selected conditions improve behavior on held-out problems.

\subsection{Task instantiations and equivalent views}
\label{sec:exposure}
\paragraph{Sudoku.}
Here $c$ contains the clues and rules, $y^\star$ is a reference board, and $s_t$ consists of visible assignments $A_t$. An offline one-wave oracle identifies frontier $F_t$ and forms
\begin{equation}
F_t=\mathcal O_{\mathrm{1wave}}(y^\star,A_t),\qquad A_{t+1}=A_t\cup F_t.
\label{eq:frontier}
\end{equation}
The implemented hybrid procedure supplies a fallback expansion when required. The retained state records define the training distribution. With $H_t=C\setminus A_t$, Eq.~\ref{eq:lts_general} becomes dense completion of the hidden cells,
\begin{equation}
\ell_{\mathrm{RSM}}(\theta;x_t,y^\star)
=\frac1{|H_t|}\sum_{i\in H_t}-\log p_{\theta,i}(y_i^\star\mid x_t).
\label{eq:dense}
\end{equation}
This is dense completion, not frontier-only supervision. Oracle rules, fallback validation, and the sampling probabilities determine the realized training distribution.

\paragraph{Maze and other instantiations.}
For maze completion, walls and start/end markers remain visible. RSM uses cached intermediate assignments and supervises every still-hidden corridor target. The experiment's state construction is distinct from random masking at an arbitrary ratio (Section~\ref{sec:maze}). Verified derivations or execution-derived values provide analogous constructions in mathematics and programming (Appendix~\ref{app:general_lts}); these illustrations are not claims of measured training transfer to those domains.

\paragraph{Presentation randomization.}
At the population-objective level, PR averages over $g\sim\operatorname{Unif}(G)$ and transports inputs and targets together. Write $gi$ for the corresponding transformed target index. The joint risk is
\begin{equation}
\mathcal L_{\mathrm{RSM+PR}}(\theta)
=\E_{e,t,g}\frac1{|H_t|}\sum_{i\in H_t}
-\log p_{\theta,gi}((gy^\star)_{gi}\mid gx_t).
\label{eq:augmented}
\end{equation}
RSM chooses conditional tasks. PR supplies equivalent representations. We use $G=D_4$ for Sudoku and maze, transporting cached states rather than rerunning the state generator. The initial recipe study materializes all eight D4 views and shuffles individual examples. The matched factorial samples a uniform view online in every PR arm. Grouped orbit averaging uses a different minibatch schedule. These procedures target the same uniformly weighted risk under matched state weights~\citep{chen2020group}, not the same finite optimization trajectory. Appendix~\ref{app:exp-posttrain} specifies the schedules and retention rules. Post-training changes parameters or adapters, not the backbone or native decoding rule.
\subsection{Conditional-state structure and supervision allocation}
\label{sec:allocation}
PR averages cross-entropy over equivalent views. The standard opinion-pool decomposition separates its consensus and disagreement components (Appendix~\ref{app:opinion}). RSM selects the state--target pairs on which this objective is evaluated. For nested reveal states, target $i$ remains supervised until reveal time $\tau(i)$, accumulating coefficient
\begin{equation}
\omega_i=\sum_{t<\tau(i)}\frac{\pi_e(t)}{|H_t|}.
\label{eq:omega}
\end{equation}
Its loss also depends on the other information visible at each occurrence. Appendix~\ref{app:matched} tests whether conditional-state structure matters beyond these marginal weights.

\subsection{The positional information compatible with grid symmetry}
\label{sec:rigidity}
Let $\Lambda_n=\{0,\ldots,n-1\}^2$, $p(r,c)=nr+c$, and $d(u,v)=p(v)-p(u)$. Consider full, unmasked attention with finite scores
\begin{equation}
L_{uv}(Q,K)=\ell_{d(u,v)}(Q_u,K_v),\qquad
\operatorname{Att}_\ell(Q,K,V)=\operatorname{softmax}_{v}(L)V.
\end{equation}
The position-only action satisfies $(\Pi_gQ)_{gu}=Q_u$.

\begin{theorem}[Serialization rigidity for relative-score attention]
\label{thm:rigidity}
For $n\ge2$, $\operatorname{Att}_\ell(\Pi_gQ,\Pi_gK,\Pi_gV)=\Pi_g\operatorname{Att}_\ell(Q,K,V)$ for every $g\in D_4$ and all $Q,K,V$ if and only if
\begin{equation}
\ell_{d(gu,gv)}(q,k)=\ell_{d(u,v)}(q,k)\quad\text{for all }g,u,v,q,k.
\label{eq:tying}
\end{equation}
The induced offset classes are $\{0\}$ and the nonzero offsets for even $n$; for odd $n$, they are $\{0\}$, nonzero even offsets, and odd offsets. In particular, if $\mathcal T_n$ is the space of scalar row-major offset kernels and $\mathcal C_n$ the space of $D_4$-invariant pair kernels, then
\begin{equation}
\dim(\mathcal T_n\cap\mathcal C_n)=\begin{cases}2,&n\text{ even},\\3,&n\text{ odd}.\end{cases}
\label{eq:dimension}
\end{equation}
\end{theorem}

The compatible positional vocabulary remains constant as the grid grows. For $n=9$, 161 offsets reduce to three classes. The reduction follows from the transitive closure of two sharing rules: symmetry-related displacements share a score, and equal serialized offsets share a score. Equivariance of attention outputs identifies the attention matrices. The self-edge removes softmax's additive ambiguity. Appendix~\ref{app:rigidity} gives the proof and the general transport-graph formulation.

For bilinear-affine scores $\ell_d(q,k)=q^\top M_dk+b_d$, Eq.~\ref{eq:tying} ties $M_d$ and $b_d$ within each class. Row-major RoPE uses $M_d=R_d$ \citep{su2024roformer}; a head with $R_1\ne R_9$ fails the $9\times9$ condition. The theorem specifies a component-level certificate under a position-only action. Learned representations and layer interactions can still yield equivariance in the full model. Our training intervention evaluates end-to-end behavior without imposing these positional ties.

\subsection{Presentation persistence under state selection and view refresh}
\label{sec:persistence}
Fix parameters $\theta$. For state sampler $a$ and puzzle $e$, let $m_{a,e}(g)=\E_{s\sim\pi_{a,e}}\nabla_\theta\ell_\theta(g\cdot z_{e,s})$ and $r_{a,e}(g)=m_{a,e}(g)-\E_hm_{a,e}(h)$. The within-puzzle persistence matrix and its trace are
\begin{equation}
 C_a=\sum_e\nu(e)^2\E_g[r_{a,e}(g)r_{a,e}(g)^\top],\qquad W_a=\operatorname{tr}C_a.
 \label{eq:persistence-matrix}
\end{equation}
They measure presentation-dependent alignment of puzzle-averaged gradients, not merely similarity of partial boards.
\begin{theorem}[Block-refresh covariance]
\label{thm:refresh-main}
Draw puzzles independently from $\nu$ and states independently conditional on puzzles. Partition $n$ draws into blocks of lengths $n_b$. Within each block reuse one independently uniform presentation per puzzle, independently of the state draws. At fixed $\theta$, for the average gradient $\widehat G_a$,
\begin{equation}
 \operatorname{Cov}(\widehat G_a)=\frac{\Sigma_a}{n}+\alpha C_a,\qquad
 \alpha=\frac{\sum_b n_b(n_b-1)}{n^2},
 \label{eq:refresh-covariance}
\end{equation}
where $\Sigma_a$ is the covariance of one gradient draw, assuming finite second moments.
\end{theorem}
Fresh PR has $\alpha=0$. A single fixed random view table has $\alpha=1-1/n$. Refresh removes the reuse component, not ordinary sampling noise. The reduction in gradient mean-squared error is $\alpha W_a$, hence its state-sampler interaction is $\alpha(W_R-W_0)$. Appendices~\ref{app:persistence-theory}--\ref{app:frozen-bank} prove these results, give the random-objective statement along training, and show why freezing a random-state bank can itself increase persistence. These are learning-signal results. The sign and magnitude of the trained model's behavioral interaction are tested, not assumed.

\section{Experiments}
\label{sec:experiments}
We separate presentation-level diagnosis, matched attribution, complete-recipe utility, and the intervention on presentation refresh. Original single-run tables are retained in the appendix alongside the additional evaluations.
\subsection{Protocol}
\label{subsec:experiments-protocol}
\paragraph{Data and models.}
Controlled 5M--100M grid-token MDLMs use DiT backbones and SUBS parameterization on \textsc{Kaggle-3M}: 500 test puzzles, 2,000 validation puzzles, 80,000 reserved for post-training, and the remaining approximately 2.9 million for pretraining. AC$k$ denotes $k$ forced target cells. The Sudoku post-training studies score full completion of the original blanks on the common AC24 test inputs. We evaluate LLaDA-8B, Dream-7B, iLLaDA-8B, and LLaDA-MoE on Sudoku, 4,318 HumanEval-Infilling pairs~\citep{chen2021evaluating,bavarian2022efficienttraininglanguagemodels}, and admitted MATH-500 pairs~\citep{hendrycks2021measuring,lightman2023letsverifystepstep}. Appendices~\ref{app:exp-details} and~\ref{app:ar-ladder} give the protocols and AR baselines.
\paragraph{Evaluation and uncertainty.}
Outputs are canonicalized before scoring. Parsing failures remain incorrect. Original large-model results use the reported greedy protocol. New same-input and cross-presentation controls use identical settings within each comparison. Symbolic evaluations use $K=16$ samples per view, temperature 1, and 128 steps. We retain sampled extrema and sECR as diagnostics, and add Eq.~\ref{eq:unbiased-v}. Unless marked crossed, intervals resample paired base puzzles conditional on the trained runs. Crossed intervals resample seed blocks and puzzles. Fine-tuning repetitions are conditional on the specified pretrained initialization, not independent pretraining runs.
\subsection{Presentation-dependent success beyond same-input variability}
\label{subsec:experiments-large-dllms}

\begin{table}[t]
\centering\small
\caption{\textbf{Correctness across equivalent presentations.} Percentages from the original reported greedy evaluations. New repeat and sampled controls are separate in Appendix~\ref{app:decoding-controls}. Sudoku uses eight views; code and mathematics use original--transformed pairs. MATH-500 pools three transformations. cLE is $100\,\mathrm{ALL}/\mathrm{ANY}$.}
\label{tab:large-dllm}
\setlength{\tabcolsep}{4pt}
\begin{tabular}{lllrrrr}
\toprule
Task & Model & Setting & Acc & ANY & ALL & cLE\\
\midrule
Sudoku & LLaDA-8B & AC16 & 49.7 & 97.7 & 1.8 & 1.8\\
 & LLaDA-8B & AC24 & 20.3 & 75.2 & 0.1 & 0.1\\
 & LLaDA-8B & AC32 & 3.9 & 24.9 & 0.0 & 0.0\\
 & Dream-7B & AC24 & 89.3 & 100.0 & 53.8 & 53.8\\
 & Dream-7B & AC32 & 68.0 & 98.9 & 17.4 & 17.6\\
 & iLLaDA-8B & AC32 & 75.2 & 99.0 & 23.8 & 24.0\\
 & LLaDA-MoE & AC32 & 66.3 & 98.9 & 12.9 & 13.0\\
\midrule
Code infilling & LLaDA-8B & Pairs & 43.5 & 57.4 & 29.5 & 51.4\\
 & Dream-7B & Pairs & 46.8 & 63.2 & 30.3 & 47.9\\
 & iLLaDA-8B & Pairs & 30.2 & 44.1 & 16.2 & 36.7\\
 & LLaDA-MoE & Pairs & 43.6 & 60.8 & 26.4 & 43.4\\
\midrule
MATH-500 & LLaDA-8B & Pairs & 29.0 & 38.7 & 19.4 & 50.1\\
 & Dream-7B & Pairs & 18.4 & 24.9 & 11.9 & 47.8\\
 & iLLaDA-8B & Pairs & 26.3 & 34.5 & 18.1 & 52.5\\
 & LLaDA-MoE & Pairs & 34.4 & 47.4 & 21.3 & 44.9\\
\bottomrule
\end{tabular}
\end{table}

LLaDA-8B solves 97.7\% of AC16 states in at least one presentation but 1.8\% in all eight. The other models show similarly large recorded gaps (Table~\ref{tab:large-dllm}). In the paired code and MATH evaluations, the transformations preserve the answer, yet correctness changes between presentations (Appendix~\ref{app:per-transform}).

These recorded gaps come from greedy decoding, so each flip is caused by the change of presentation. To test whether the gaps also persist under sampling, we compare two ways of decoding a problem with $m$ presentations. A pseudo-orbit consists of $m$ independent decodes of its original presentation, and a real orbit consists of one decode per presentation. Under equal view success probabilities and independent decoding, their ANY/ALL distributions coincide. In all 15 settings, the gap is larger for the real orbit: by 21.8--81.5 points on Sudoku, 9.5--15.0 on code, and 3.1--6.8 on MATH. Since both orbits contain the same sampling noise, this additional gap reflects presentation. Appendix~\ref{app:decoding-controls} gives the full results.

\subsection{State selection changes the effect of presentation randomization}
\label{subsec:experiments-posttraining}
\paragraph{Matched attribution.}
All arms share puzzle frequencies and hidden counts. A/B use random states without/with online PR, C/D use reasoning states, and F/E additionally match each random target's expected supervision coefficient to RSM. Adding F completes the weight-matched factorial $I^{\rm wt}_M=(M_D-M_C)-(M_E-M_F)$. The 50M study has five fine-tuning seeds per arm and the 100M study three (Appendix~\ref{app:matched}).

\begin{table}[t]
\centering\small
\caption{\textbf{Matched state--presentation interactions.}
Differences in percentage points with 95\% puzzle-bootstrap intervals conditional on trained runs. The original A--E results and augmented endpoints are retained in Appendix~\ref{app:matched}.}
\label{tab:matched-main}
\begin{tabular}{lcc}
\toprule
Contrast & 50M (5 seeds) & 100M (3 seeds)\\
\midrule
$I_{\rm mean}=(D-C)-(B-A)$ & $+1.57\ [1.09,2.07]$ & $+2.27\ [1.72,2.84]$\\
$I^{\rm wt}_{\rm mean}$ & $+1.51\ [1.04,1.98]$ & $+2.45\ [1.89,3.01]$\\
$I^{\rm wt}_{\rm sECR}$ & $+4.06\ [2.61,5.56]$ & $+6.32\ [4.70,7.92]$\\
\bottomrule
\end{tabular}
\end{table}
The interaction appears in mean success itself, not only in an extrema ratio: at 100M, PR changes mean success by $-1.00\ [-1.37,-0.64]$ points on weight-matched random states but $+1.44\ [0.84,2.04]$ on reasoning states. Endpoint utility is a separate question. D--B mean success is $+0.94\ [0.57,1.29]$ at 50M and $+0.32\ [-0.03,0.68]$ at 100M, and sECR shows no clear endpoint advantage (Appendix~\ref{app:matched}).

\paragraph{The result is not confined to sampled extrema.}
The unbiased dispersion-reduction interaction $\widehat J$ is $+0.00289\ [0.00185,0.00394]$ at 50M and $+0.00440\ [0.00338,0.00548]$ at 100M. Independently of the choice of extrema, the 50M five-seed token-marginal analysis retains an interaction of $-0.00434$ nats after whole-output repartition calibration, with crossed interval $[-0.00547,-0.00322]$. All five seed-specific contrasts are negative. Lower leakage is better. This corroborates a stronger PR reduction under RSM, not a lower D endpoint than B or equality of full output laws (Appendix~\ref{app:measurement-results}).

\paragraph{Complete recipes reproduce across fine-tuning seeds.}
The initial recipe comparison changes puzzle retention and masking as well as state structure. It is distinct from attribution. We repeat its four main recipes with three fine-tuning seeds, evaluating all twelve checkpoints in one batch-64 harness (Table~\ref{tab:combined_post_training}). RSM+PR ranks first in all three summaries in every seed. Relative to canonical SFT, its mean-success gain is $+3.70\ [3.19,4.22]$ points under crossed resampling. Relative to PR, it is $+4.84\ [4.30,5.40]$. The factorial mean-success interaction is $+2.09\ [1.52,2.69]$. Original single-run baselines and all contrasts remain in Appendix~\ref{app:posttrain-ci}.
\begin{table}[t]
\centering\small
\caption{\textbf{Three-seed 50M recipe repetition.} Mean (between-seed SD), in percent. Each seed uses the same pretrained initialization and 9,000 updates. All checkpoints use one common evaluator; these are not the original sampled outputs.}
\label{tab:combined_post_training}
\begin{tabular}{lrrr}
\toprule
Recipe & MeanSucc & Estimated $R_{\min}$ & sECR\\
\midrule
Canonical & 42.81 (0.10) & 26.40 (0.16) & 44.38 (0.22)\\
PR & 41.68 (0.12) & 25.94 (0.11) & 44.94 (0.17)\\
RSM & 45.56 (0.04) & 25.56 (0.10) & 38.97 (0.17)\\
RSM+PR & \textbf{46.52 (0.05)} & \textbf{29.79 (0.07)} & \textbf{47.15 (0.08)}\\
\bottomrule
\end{tabular}
\end{table}

\subsection{Intervening on presentation persistence}
\label{sec:mechanism-experiments}
At the common pretrained 50M checkpoint, the projected persistence trace in the last two blocks and head is approximately 11 times larger for reasoning states than for matched online random states. The raw difference is $1.78\times10^{-4}\ [1.51\times10^{-4},2.07\times10^{-4}]$. Separately trained three-seed models change only the presentation-refresh schedule. Their sECR interaction increases from 0.06 (fixed) to 2.44 and 3.13 (intermediate refresh) and 4.30 points (fresh). The intermediate difference $0.69\ [-0.51,1.92]$ is unresolved. A frozen nested-random trajectory control, without a reasoning solver, gives a smaller positive interaction $+1.74\ [0.66,2.90]$. Appendix~\ref{app:mechanism-results} reports the tested alternatives, including non-confirmed predictions. The interventions support a state-dependent reuse mechanism. The covariance identity alone does not predict final-model effect sizes.

\subsection{Matched mazes and the 8B recipe}
\label{sec:maze}\label{subsec:experiments-transfer}
\paragraph{Mazes.}
On 500 held-out n15 mazes, the new 100M comparison matches hidden-cell density, marginal target weights, and online PR, with three fine-tuning seeds per arm. RSM+PR raises mean success from 92.0\% to 94.1\%, a crossed difference of $+2.10\ [1.58,2.67]$ points. Its sECR difference is unresolved, $+0.74\ [-0.17,1.63]$. This establishes a solving advantage, not a normalized-stability advantage. The original random-ratio recipe and initialization sweeps are retained separately in Appendix~\ref{app:maze}.
\paragraph{LLaDA-8B.}
The LoRA recipe comparison raises ALL from 47.7\% under canonical SFT to 74.8\% under RSM and 86.1\% under RSM+PR, while PR alone gives 46.0\%. A sampled control on the same checkpoints gives same-input gaps of 0.0--0.2 points and real-orbit gaps of 51.6/50.2/25.2/12.8 for Canonical/PR/RSM/RSM+PR. A density-matched comparison uses random states with the same puzzles, hidden-cell counts, and number of up-weighted targets as the RSM states, under the identical LoRA protocol. RSM+PR exceeds matched random+PR by 33.5 points in all-view correctness, and RSM exceeds matched random states by 27.8 points (Appendix~\ref{app:8b}).
\subsection{Inference-time alternatives and autoregressive scope}
\label{sec:baselines}
Applying the same inference procedure to every recipe separates training from query budget. Eight-call verify-and-select is strong (about 92\% mean success), yet RSM+PR improves joint success over Canonical by $+4.40\ [1.20,7.60]$ points and over PR by $+7.40\ [4.00,11.00]$. RSM alone has the highest joint point estimate, 73.8\% versus 72.6\%. Cell-wise voting lowers mean success in all four recipes. Uniform random-view decoding symmetrizes the output law while preserving orbit-mean success, so post-training is not necessary for distributional symmetrization (Appendices~\ref{app:symmetrization} and~\ref{app:inference}).

Greedy Qwen2.5-7B-Instruct and Llama-3.1-8B-Instruct also change correctness under answer-preserving Sudoku and MATH presentations. Parsing failures do not explain all mixed-correctness puzzles. In MATH, base-problem-balanced ANY/ALL are 47.0/29.9 and 43.9/22.6, respectively (Appendix~\ref{app:ar-ladder}). Thus presentation dependence is not diffusion-exclusive. Its presence in bidirectional masked diffusion shows that causal masking is not necessary. Choosing generation order within a serialization does not guarantee stability across serializations.

\section{Limitations and Discussion}
\label{sec:discussion_and_limitations}
Our training interventions are evaluated on grid-structured tasks, where exact symmetry groups make orbit evaluation precise. The code and MATH evaluations show that presentation dependence extends further, and applying RSM there with verified intermediate facts or execution traces is a natural next step. Presentation dependence also appears in autoregressive models. What masked diffusion adds is an explicit training lever, the choice of supervised partial states, which shapes how presentation exposure acts. Finally, the rigidity theorem concerns a single attention component, and the positional pilot in Appendix~\ref{app:positional-pilot} suggests that symmetry-aware positional indexing is a promising complement to RSM+PR.

\section{Conclusion}
\label{sec:conclusion}
Equivalent presentations expose changes in correctness that aggregate accuracy can conceal. Exact transported targets, decoding controls, and finite-sample population measurement establish this effect without relying solely on sampled extrema. Matched experiments show that conditional-state selection changes what presentation augmentation accomplishes. Presentation-persistence theory and refresh interventions connect that interaction to the learning signal, while direct endpoint comparisons identify where it improves solving. The aim is both to solve more problems and to retain those successes when their presentation changes.

\bibliography{references}
\bibliographystyle{le_preprint}
\clearpage
\appendix

\section{Additional Proofs and Protocol Details}
\label{app:method-appendix}

All logarithms are natural.  The group size is denoted by $m=|G|$.  The principal grid evaluation uses $G=D_4$ and $m=8$; the additional-group experiments specify their own transformation families.  The output space includes the invalid parse
symbol $\bot$, with $g\bot=\bot$.  All probability statements below are for a
fixed base logical state $x$ unless stated otherwise.

\subsection{Forced-target validity of SUDOKU-AC-ORBIT}
\label{app:forced-target}\label{app:forced}

This appendix proves the claim used in Section~\ref{sec:le}:
if the target cells are forced in the canonical board, then the corresponding
target cells are forced in every rotated or reflected view.  Thus every view has the same transported target and an isomorphic constraint system.

Let $\Lambda=\{1,\ldots,9\}^2$ be the set of Sudoku cells.  A completed Sudoku
solution is a function $s:\Lambda\to\{1,\ldots,9\}$ satisfying the row, column,
and $3\times3$ box constraints.  For a set of anchors $A$, let $\mathcal S(A)$ be
the set of valid completed Sudoku grids that agree with the anchored digits.
For $U\subseteq\Lambda$, write $s_U$ for the restriction of $s$ to $U$.

\begin{proposition}[Forced targets are preserved under $D_4$]
\label{prop:forced-target-d4}
Let $g\in D_4$.  Then
\begin{equation}
    \mathcal S(gA)=g\mathcal S(A):=\{gs:s\in\mathcal S(A)\}.
\end{equation}
Consequently, if
\begin{equation}
    \{s_U:s\in\mathcal S(A)\}=\{s_U^\star\},
\end{equation}
then
\begin{equation}
    \{t_{gU}:t\in\mathcal S(gA)\}=\{(gs^\star)_{gU}\}.
\end{equation}
\end{proposition}

\begin{proof}
Every element of $D_4$ is a rotation or reflection of the $9\times9$ square.  It
maps rows to rows or columns, columns to columns or rows, and $3\times3$ boxes to
$3\times3$ boxes.  Thus $s$ satisfies the Sudoku constraints if and only if $gs$
satisfies the Sudoku constraints.  Also, $s$ agrees with anchors $A$ if and only
if $gs$ agrees with transformed anchors $gA$.  Hence $\mathcal S(gA)=g\mathcal
S(A)$.

If $t\in\mathcal S(gA)$, then by the first part there exists $s\in\mathcal S(A)$
with $t=gs$.  Therefore $t_{gU}=(gs)_{gU}=g(s_U)$.  Since $s_U=s_U^\star$ for all
$s\in\mathcal S(A)$, the transformed restriction is always $(gs^\star)_{gU}$.
This proves target uniqueness in every view.
\end{proof}

\subsection{View leakage and the output view spectrum}
\label{app:view-spectrum}

We connect distributional logical equivariance to view leakage and its spectrum. Zero view leakage is exactly distributional logical equivariance on the orbit of $x$, and the
$\chi^2$-version of view leakage decomposes exactly into the nontrivial Fourier
modes of the group.  The KL leakage $L_{\mathrm{out}}$ is not itself an exact
Fourier norm; it is equivalent to that Fourier norm up to constants depending
only on $|G|$.

Sample $V\sim\mathrm{Unif}(G)$, generate $Y\sim K_\theta(\cdot\mid Vx)$, and
canonicalize $Y^c=V^{-1}Y$.  Let
\begin{equation}
    P_g := g^{-1}_{\#}K_\theta(\cdot\mid gx),
    \qquad
    \overline P := \frac1m\sum_{g\in G}P_g.
\end{equation}
Then the conditional law of $Y^c$ given $V=g$ is $P_g$, and the marginal law of
$Y^c$ is $\overline P$.

\begin{proposition}[Zero leakage is distributional LE]
\label{prop:zero-leakage-le}
\begin{equation}
    L_{\mathrm{out}}(x):=I(V;Y^c)
    =\frac1m\sum_{g\in G}D_{\mathrm{KL}}(P_g\Vert\overline P).
\end{equation}
Moreover,
\begin{equation}
    L_{\mathrm{out}}(x)=0
    \quad\Longleftrightarrow\quad
    P_g=P_h\quad\forall g,h\in G.
\end{equation}
\end{proposition}

\begin{proof}
The joint law of $(V,Y^c)$ is
\begin{equation}
    \Pr(V=g,Y^c\in B)=\frac1mP_g(B).
\end{equation}
The marginal law of $Y^c$ is $\overline P$.  Therefore
\begin{equation}
    I(V;Y^c)
    =D_{\mathrm{KL}}(\Pr_{V,Y^c}\Vert \Pr_V\otimes \Pr_{Y^c})
    =\frac1m\sum_{g\in G}D_{\mathrm{KL}}(P_g\Vert\overline P).
\end{equation}
Mutual information is zero if and only if $V$ and $Y^c$ are independent.  Since
all views have positive probability, independence is equivalent to
$P_g=\overline P$ for every $g$, which is equivalent to $P_g=P_h$ for all
$g,h\in G$.
\end{proof}

We now decompose the view dependence by irreducible representations.  Let
$\widehat G$ be a set of unitary irreducible representations of $G$.  For
$\rho\in\widehat G$, let $d_\rho$ be its dimension.  The trivial representation
is denoted by $\mathbf 1$.

Because $\overline P$ is an average containing each $P_g$, we have $P_g\ll
\overline P$.  Define
\begin{equation}
    r_g(y):=\frac{dP_g}{d\overline P}(y).
\end{equation}
Then $0\le r_g(y)\le m$ for $\overline P$-almost every $y$, and
\begin{equation}
    \frac1m\sum_{g\in G}r_g(y)=1.
\end{equation}
For each $y$, view $g\mapsto r_g(y)$ as a function on the group, and define its
Fourier coefficient by
\begin{equation}
    \widehat r_\rho(y)
    :=\frac1m\sum_{g\in G}r_g(y)\rho(g)^*.
\end{equation}

\begin{theorem}[Output view spectrum]
\label{thm:appendix-view-spectrum}
Define
\begin{equation}
    \chi_G^2(x)
    :=\int \frac1m\sum_{g\in G}\bigl(r_g(y)-1\bigr)^2\,d\overline P(y).
\end{equation}
Then
\begin{equation}
    \chi_G^2(x)
    =\sum_{\rho\ne\mathbf 1}
    d_\rho\int\|\widehat r_\rho(y)\|_{\mathrm{HS}}^2\,d\overline P(y).
    \label{eq:appendix-chi-spectrum}
\end{equation}
Furthermore, there exist constants $0<a_m\le b_m<\infty$, depending only on
$m=|G|$, such that
\begin{equation}
    a_m\chi_G^2(x)
    \le L_{\mathrm{out}}(x)
    \le b_m\chi_G^2(x).
    \label{eq:appendix-kl-chi-comparison}
\end{equation}
\end{theorem}

\begin{proof}
For each fixed $y$, apply the normalized Plancherel identity on the finite group:
\begin{equation}
    \frac1m\sum_{g\in G}|f(g)|^2
    =\sum_{\rho\in\widehat G}d_\rho\|\widehat f(\rho)\|_{\mathrm{HS}}^2,
    \qquad
    \widehat f(\rho)=\frac1m\sum_{g\in G}f(g)\rho(g)^*.
\end{equation}
Use $f(g)=r_g(y)$.  The trivial coefficient is
\begin{equation}
    \widehat r_{\mathbf 1}(y)=\frac1m\sum_{g\in G}r_g(y)=1.
\end{equation}
Thus Plancherel gives
\begin{equation}
    \frac1m\sum_{g\in G}\bigl(r_g(y)-1\bigr)^2
    =\sum_{\rho\ne\mathbf 1}d_\rho\|\widehat r_\rho(y)\|_{\mathrm{HS}}^2.
\end{equation}
Integrating over $\overline P$ proves Eq.~(\ref{eq:appendix-chi-spectrum}).

For the KL comparison, let
\begin{equation}
    \varphi(t):=t\log t-t+1,\qquad \varphi(0):=1.
\end{equation}
Since $\int r_g\,d\overline P=1$,
\begin{equation}
    D_{\mathrm{KL}}(P_g\Vert\overline P)
    =\int r_g\log r_g\,d\overline P
    =\int \varphi(r_g)\,d\overline P.
\end{equation}
The ratio $\varphi(t)/(t-1)^2$, with value $1/2$ at $t=1$ by continuous
extension, is positive and finite on the compact interval $[0,m]$.  Hence there
are constants $a_m,b_m$ such that
\begin{equation}
    a_m(t-1)^2\le \varphi(t)\le b_m(t-1)^2
    \qquad\forall t\in[0,m].
\end{equation}
Apply this pointwise to $t=r_g(y)$, average over $g$, and integrate over
$\overline P$.
\end{proof}

The following proposition connects the spectrum to the correctness metrics.  It
is useful because it turns an observed Any--All gap into a lower bound on
nontrivial view information.

\begin{proposition}[Orbit failure forces view leakage]
\label{prop:any-all-forces-leakage}
For deterministic decoding,
\begin{equation}
    L_{\mathrm{out}}(x)
    \ge h_2(1/m)\bigl(\mathrm{Any}(x)-\mathrm{All}(x)\bigr),
\end{equation}
where $h_2(p)=-p\log p-(1-p)\log(1-p)$.
For stochastic decoding, if $p_g(x)=P_g(\{y^\star(x)\})$, then
\begin{equation}
    L_{\mathrm{out}}(x)
    \ge \frac{(C_{\max}(x)-R_{\min}(x))^2}{m}.
\end{equation}
\end{proposition}

\begin{proof}
For deterministic decoding, $Y^c=\widehat y_V(x)$ is a deterministic function of
$V$, so $L_{\mathrm{out}}(x)=H(Y^c)$.  If
$\mathrm{Any}(x)-\mathrm{All}(x)=0$, the stated inequality is immediate.  If the
gap is one, then one canonicalized prediction equals the unique target and at
least one does not.  Hence $Y^c$ is a nonconstant function of a uniform
$m$-point variable.  The minimum entropy of such a nonconstant image occurs when
one fiber has size $1$ and another has size $m-1$, giving $H(Y^c)\ge h_2(1/m)$.

For stochastic decoding, let $C=\mathbf 1[Y^c=y^\star(x)]$.  Data processing
gives $I(V;Y^c)\ge I(V;C)$.  With $\overline p=m^{-1}\sum_g p_g$,
\begin{equation}
    I(V;C)=\frac1m\sum_{g\in G}\mathrm{kl}(p_g\Vert\overline p).
\end{equation}
Pinsker's inequality for Bernoulli laws gives
$\mathrm{kl}(p_g\Vert\overline p)\ge 2(p_g-\overline p)^2$.  Let
$\delta=C_{\max}-R_{\min}$.  Choose views $g_+,g_-$ attaining the maximum and
minimum.  Then
\begin{equation}
    |p_{g_+}-\overline p|+|p_{g_-}-\overline p|\ge \delta,
\end{equation}
and therefore
\begin{equation}
    (p_{g_+}-\overline p)^2+(p_{g_-}-\overline p)^2\ge \delta^2/2.
\end{equation}
Thus
\begin{equation}
    L_{\mathrm{out}}(x)
    \ge I(V;C)
    \ge \frac2m\sum_g(p_g-\overline p)^2
    \ge \frac{\delta^2}{m}.
\end{equation}
\end{proof}

\subsection{Row-major Toeplitz--\texorpdfstring{$D_4$}{D4} rigidity}
\label{app:toeplitz-d4}

We first establish the scalar offset-class result, then use it to prove the attention-head theorem in Section~\ref{sec:rigidity}.

Let
\begin{equation}
    \Lambda_n:=\{0,\ldots,n-1\}^2,
    \qquad
    p(r,c):=nr+c.
\end{equation}
A scalar row-major relative-position kernel has the form
\begin{equation}
    B_{uv}=a_{p(v)-p(u)}.
\end{equation}
Let $\mathcal T_n$ be the vector space of all such kernels.  Let
\begin{equation}
    \mathcal C_n:=\{B:B_{gu,gv}=B_{uv}\ \forall g\in D_4,\ u,v\in\Lambda_n\}
\end{equation}
be the space of $D_4$-invariant scalar pair kernels.

\begin{theorem}[Row-major Toeplitz--$D_4$ rigidity]
\label{thm:appendix-toeplitz-d4}
For $n\ge2$,
\begin{equation}
    \dim(\mathcal T_n\cap\mathcal C_n)
    =
    \begin{cases}
    2, & n\text{ even},\\
    3, & n\text{ odd}.
    \end{cases}
\end{equation}
More explicitly, if $n$ is even, then every $B\in\mathcal T_n\cap\mathcal C_n$
has the form
\begin{equation}
    B_{uv}=\alpha\mathbf 1[p(v)-p(u)=0]
    +\beta\mathbf 1[p(v)-p(u)\ne0].
\end{equation}
If $n$ is odd, then every $B\in\mathcal T_n\cap\mathcal C_n$ has the form
\begin{equation}
    B_{uv}
    =\alpha\mathbf 1[d=0]
    +\beta\mathbf 1[d\ne0\text{ and }d\text{ even}]
    +\gamma\mathbf 1[d\text{ odd}],
    \qquad d=p(v)-p(u).
\end{equation}
In particular, for the $9\times9$ Sudoku grid, the certified row-major scalar
relative-position algebra has only the three classes $0$, nonzero-even, and odd.
\end{theorem}

\begin{proof}
For a pair of cells $u=(r,c)$ and $v=(r',c')$, define the displacement
\begin{equation}
    (x,y):=(r'-r,c'-c),
    \qquad
    x,y\in\{-(n-1),\ldots,n-1\}.
\end{equation}
The row-major offset is
\begin{equation}
    \delta(x,y):=p(v)-p(u)=nx+y.
\end{equation}
The group $D_4$ acts on displacements by sign changes and coordinate swaps:
\begin{equation}
    (x,y)\mapsto(\pm x,\pm y),
    \qquad
    (x,y)\mapsto(\pm y,\pm x),
\end{equation}
with the signs constrained by the chosen rotation or reflection.

The only subtle point is that the same offset may have several feasible
displacement representations.  For example, $nx+y=nx'+y'$ can hold with
$(x,y)\ne(x',y')$.  Define an equivalence relation $\sim$ on offsets by the
following generators: two offsets are equivalent if they are the same integer, or
if they arise from feasible displacements related by a $D_4$ transformation.
Then $B\in\mathcal T_n\cap\mathcal C_n$ exactly means $a_d=a_{d'}$ whenever
$d\sim d'$.  Thus the dimension of $\mathcal T_n\cap\mathcal C_n$ is the number
of equivalence classes of offsets.

First, offset $0$ is a singleton class.  If $nx+y=0$ with
$|x|,|y|\le n-1$, then $x\ne0$ would imply $|nx|\ge n>|y|$, impossible.  Hence
$x=y=0$.

Second, every nonzero offset is equivalent to one of $1,\ldots,n-1$.  By a
$180^\circ$ rotation, $d\sim -d$, so take $d>0$.  Write
$d=nq+r$ with $0\le r\le n-1$.  If $q=0$, then $d=r\in\{1,\ldots,n-1\}$.  If
$q\ge1$ and $r=0$, then
\begin{equation}
    d=\delta(q,0)\sim\delta(0,q)=q<d.
\end{equation}
If $q\ge1$ and $r>0$, then
\begin{equation}
    d=\delta(q,r)\sim\delta(q,-r)=nq-r<d.
\end{equation}
Repeating this strict descent reaches a positive offset in $\{1,\ldots,n-1\}$.

Third, for every $1\le k\le n-3$,
\begin{equation}
    k\sim k+2.
    \label{eq:k-kplus2}
\end{equation}
Indeed,
\begin{align}
    k
    &=\delta(1,k-n) \\
    &\sim \delta(n-k,-1) \\
    &= n(n-k)-1 \\
    &= \delta(n-k-1,n-1) \\
    &\sim \delta(-(n-k-1),n-1) \\
    &= \delta(-(n-k-2),-1) \\
    &\sim \delta(1,k+2-n) \\
    &= k+2.
\end{align}
All displacements used above lie in $[-(n-1),n-1]^2$ when $1\le k\le n-3$.

Now suppose $n$ is odd.  Since
\begin{equation}
    \delta(x,y)=nx+y\equiv x+y\pmod 2,
\end{equation}
and $D_4$ only changes signs and swaps coordinates, offset parity is invariant.
By Eq.~(\ref{eq:k-kplus2}), all positive small odd offsets are equivalent and all
positive small even offsets are equivalent.  The descent argument then shows
that all nonzero odd offsets form one class and all nonzero even offsets form
one class.  Together with the singleton class $\{0\}$, this gives three classes.

Finally suppose $n$ is even.  For $n=2$, direct inspection gives
$1=\delta(0,1)\sim\delta(1,0)=2$, so all nonzero offsets are equivalent.  For
$n\ge4$,
\begin{equation}
    1=\delta(1,1-n)
    \sim \delta(1,n-1)
    =2n-1
    =\delta(2,-1)
    \sim \delta(1,-2)
    =n-2.
\end{equation}
The offset $n-2$ is even.  By Eq.~(\ref{eq:k-kplus2}), it is equivalent to $2$.
Hence $1\sim2$, so the odd and even nonzero classes merge.  The descent argument
then shows that all nonzero offsets form one class.  Together with $\{0\}$, this
gives two classes.
\end{proof}

\paragraph{Consequence for 1D RoPE.}
Let $M_d$ be the matrix used by a 1D relative positional component at row-major
offset $d$.  For fixed query/key contents $a,b$, the scalar positional term has
form
\begin{equation}
    B^{a,b}_{uv}=a^*M_{p(v)-p(u)}b.
\end{equation}
If this scalar term is $D_4$-invariant for every $a,b$, then Theorem
\ref{thm:appendix-toeplitz-d4} forces $M_d=M_{d'}$ whenever $d$ and $d'$ lie in
the same offset class.  On a $9\times9$ grid, this means all nonzero odd offsets
must share the same matrix, including $1$, $9$, and $-9$.  Standard nondegenerate
1D RoPE separates these offsets.  Therefore row-major 1D RoPE does not provide a
component-level scalar $D_4$ positional certificate.

This is a component-level constraint on the positional matrices.

\subsection{Row-major relative-score attention rigidity}
\label{app:relative-score-rigidity}\label{app:rigidity}

We now establish Theorem~\ref{thm:rigidity} for full, unmasked relative-score attention.

Let
\[
    \Lambda_n:=\{0,\ldots,n-1\}^2,
    \qquad
    p(r,c):=nr+c,
\]
and define the row-major offset
\[
    d(u,v):=p(v)-p(u).
\]
Let \(D_4\) act on \(\Lambda_n\), and let \(\Pi_g\) denote the induced
position-only permutation, with the convention
\[
    (\Pi_g Z)_{gu}=Z_u.
\]
The action permutes positions only and does not transform hidden channels.

We first record the offset-class fact used in the main theorem.

\begin{lemma}[Row-major Toeplitz--\(D_4\) offset classes]
\label{lem:row-major-d4-offset-classes}
Let \(\sim\) be the equivalence relation on row-major offsets generated by
\[
    d(u,v)\sim d(gu,gv)
    \qquad
    \text{for all }g\in D_4,\ u,v\in\Lambda_n .
\]
Then, for \(n\ge2\), the offset \(0\) is a singleton class. For nonzero
offsets:
\[
    \begin{cases}
    \text{all nonzero offsets form one class}, & n\text{ even},\\
    \text{all nonzero-even offsets form one class and all odd offsets form one class},
    & n\text{ odd}.
    \end{cases}
\]
In particular, for \(n=9\), any row-major offset-indexed object that is
constant on \(D_4\)-related cell pairs must have the form
\[
    a_d =
    \begin{cases}
    a_0, & d=0,\\
    a_E, & d\neq 0,\ d\equiv 0 \pmod 2,\\
    a_O, & d\equiv 1 \pmod 2.
    \end{cases}
\]
\end{lemma}

\begin{proof}
The equivalence relation is exactly the one generated by the feasible displacement transformations in Theorem~\ref{thm:appendix-toeplitz-d4}. Its proof establishes the singleton zero class, descent to offsets $1,\ldots,n-1$, parity preservation for odd $n$, and merging of the two nonzero parity classes for even $n$.
\end{proof}

We now prove the attention-head rigidity theorem from the main text.

\begin{proof}[Proof of Theorem~\ref{thm:rigidity}]
Let
\[
    \Delta_n:=\{d(u,v):u,v\in\Lambda_n\}.
\]
For each \(d\in\Delta_n\), let
\[
    \ell_d:\mathbb{R}^{r_q}\times\mathbb{R}^{r_k}\to\mathbb{R}
\]
be a finite scalar scoring function. The attention logits are
\[
    L_{uv}(Q,K)=\ell_{d(u,v)}(Q_u,K_v),
\]
and the row-stochastic attention matrix is
\[
    A_\ell(Q,K)=\operatorname{softmax}_{v}(L(Q,K)).
\]
Thus
\[
    \operatorname{Att}_\ell(Q,K,V)=A_\ell(Q,K)V.
\]

Assume first that the attention head is exactly \(D_4\)-equivariant under the
position-only action. Then, for every \(g\in D_4\) and every value array \(V\),
\[
    A_\ell(\Pi_gQ,\Pi_gK)\Pi_g V
    =
    \Pi_g A_\ell(Q,K)V.
\]
Since this identity holds for all \(V\), we may write \(W=\Pi_gV\) and obtain
\[
    A_\ell(\Pi_gQ,\Pi_gK)W
    =
    \Pi_g A_\ell(Q,K)\Pi_g^{-1}W
    \qquad
    \forall W.
\]
Therefore,
\begin{equation}
    A_\ell(\Pi_gQ,\Pi_gK)
    =
    \Pi_g A_\ell(Q,K)\Pi_g^{-1}.
    \label{eq:appendix-attn-conjugacy}
\end{equation}
Equivalently, for every \(u,v\in\Lambda_n\),
\begin{equation}
    A_\ell(\Pi_gQ,\Pi_gK)_{gu,gv}
    =
    A_\ell(Q,K)_{u,v}.
    \label{eq:appendix-attn-entrywise-conjugacy}
\end{equation}

For finite logits, equality of two row-wise softmax distributions is equivalent
to equality of the corresponding logits up to an additive constant depending
only on the row. Hence, for each fixed \(g,u,Q,K\), there exists a scalar
\(c_{g,u}(Q,K)\) such that
\begin{equation}
    L_{gu,gv}(\Pi_gQ,\Pi_gK)
    =
    L_{u,v}(Q,K)+c_{g,u}(Q,K)
    \qquad
    \forall v\in\Lambda_n.
    \label{eq:appendix-logit-row-constant}
\end{equation}
Using the row-major relative-score form and the position-only action,
\[
    (\Pi_gQ)_{gu}=Q_u,
    \qquad
    (\Pi_gK)_{gv}=K_v,
\]
so Eq.~(\ref{eq:appendix-logit-row-constant}) becomes
\begin{equation}
    \ell_{d(gu,gv)}(Q_u,K_v)
    =
    \ell_{d(u,v)}(Q_u,K_v)+c_{g,u}(Q,K)
    \qquad
    \forall v.
    \label{eq:appendix-score-plus-constant}
\end{equation}

Because the attention head is full and unmasked, the self-edge \(v=u\) is
available. Setting \(v=u\), we have
\[
    d(u,u)=d(gu,gu)=0.
\]
Therefore Eq.~(\ref{eq:appendix-score-plus-constant}) gives
\[
    \ell_0(Q_u,K_u)
    =
    \ell_0(Q_u,K_u)+c_{g,u}(Q,K),
\]
and hence
\[
    c_{g,u}(Q,K)=0.
\]
Substituting this back into Eq.~(\ref{eq:appendix-score-plus-constant}),
\[
    \ell_{d(gu,gv)}(Q_u,K_v)
    =
    \ell_{d(u,v)}(Q_u,K_v)
    \qquad
    \forall g,u,v,Q,K.
\]
Since the theorem assumes equivariance for all \(Q,K\), the vectors \(Q_u\) and
\(K_v\) can be varied arbitrarily. Thus
\[
    \ell_{d(gu,gv)}(q,k)
    =
    \ell_{d(u,v)}(q,k)
\]
for every \(g\in D_4\), every \(u,v\in\Lambda_n\), and all
\(q\in\mathbb{R}^{r_q}\), \(k\in\mathbb{R}^{r_k}\).

Conversely, suppose the score functions are tied on all \(D_4\)-related
row-major offsets:
\[
    \ell_{d(gu,gv)}(q,k)
    =
    \ell_{d(u,v)}(q,k)
    \qquad
    \forall g,u,v,q,k.
\]
Then
\begin{align}
    L_{gu,gv}(\Pi_gQ,\Pi_gK)
    &=
    \ell_{d(gu,gv)}((\Pi_gQ)_{gu},(\Pi_gK)_{gv}) \\
    &=
    \ell_{d(gu,gv)}(Q_u,K_v) \\
    &=
    \ell_{d(u,v)}(Q_u,K_v) \\
    &=
    L_{u,v}(Q,K).
\end{align}
Hence
\[
    L(\Pi_gQ,\Pi_gK)=\Pi_gL(Q,K)\Pi_g^{-1}.
\]
Row-wise softmax commutes with simultaneous row and column permutation, so
\[
    A_\ell(\Pi_gQ,\Pi_gK)
    =
    \Pi_g A_\ell(Q,K)\Pi_g^{-1}.
\]
Multiplying by \(\Pi_gV\), we get
\begin{align}
    \operatorname{Att}_\ell(\Pi_gQ,\Pi_gK,\Pi_gV)
    &=
    A_\ell(\Pi_gQ,\Pi_gK)\Pi_gV \\
    &=
    \Pi_g A_\ell(Q,K)V \\
    &=
    \Pi_g \operatorname{Att}_\ell(Q,K,V).
\end{align}
This proves the equivalence.

For \(n=9\), Lemma~\ref{lem:row-major-d4-offset-classes} gives exactly three
row-major offset classes:
\[
    d=0,
    \qquad
    d\neq0,\ d\equiv0\pmod2,
    \qquad
    d\equiv1\pmod2.
\]
Therefore the score functions must collapse as
\[
    \ell_d =
    \begin{cases}
    \ell_0, & d=0,\\
    \ell_E, & d\neq 0,\ d\equiv 0 \pmod 2,\\
    \ell_O, & d\equiv 1 \pmod 2.
    \end{cases}
\]
In particular, the offsets \(1\), \(9\), and \(-9\) are all odd offsets on a
\(9\times9\) grid, so an exactly \(D_4\)-equivariant row-major relative-score
attention head must tie their score functions.
\end{proof}

\begin{corollary}[Bilinear relative scores and RoPE]
\label{cor:bilinear-rope-collapse}
For $\ell_d(q,k)=q^\top M_dk+b_d$, exact equivariance under the assumptions of Theorem~\ref{thm:rigidity} is equivalent to tying $M_d$ and $b_d$ within the theorem's offset classes. In particular, row-major RoPE with $R_1\neq R_9$ does not satisfy the $9\times9$ tying condition.
\end{corollary}

\begin{proof}[Proof of Corollary~\ref{cor:bilinear-rope-collapse}]
Assume the relative score has the bilinear-affine form
\[
    \ell_d(q,k)=q^\top M_d k+b_d.
\]
By Theorem~\ref{thm:rigidity}, exact \(D_4\)-equivariance implies
\[
    q^\top M_{d(gu,gv)}k+b_{d(gu,gv)}
    =
    q^\top M_{d(u,v)}k+b_{d(u,v)}
\]
for every \(g\in D_4\), every \(u,v\in\Lambda_n\), and all \(q,k\).

Setting \(q=0\) gives
\[
    b_{d(gu,gv)}=b_{d(u,v)}.
\]
After subtracting the equal biases, we obtain
\[
    q^\top
    \left(M_{d(gu,gv)}-M_{d(u,v)}\right)
    k
    =
    0
    \qquad
    \forall q,k.
\]
Choosing \(q\) and \(k\) to range over standard basis vectors gives
\[
    M_{d(gu,gv)}=M_{d(u,v)}.
\]
For \(n=9\), Lemma~\ref{lem:row-major-d4-offset-classes} then implies
\[
    (M_d,b_d)=
    \begin{cases}
    (M_0,b_0), & d=0,\\
    (M_E,b_E), & d\neq 0,\ d\equiv 0 \pmod 2,\\
    (M_O,b_O), & d\equiv 1 \pmod 2.
    \end{cases}
\]

For row-major RoPE, let \(R_t\) denote the block-diagonal rotary matrix at
sequence position \(t\), with the usual property
\[
    R_a^\top R_b=R_{b-a}.
\]
Applying RoPE to query and key vectors at row-major positions \(p(u)\) and
\(p(v)\) gives, up to the conventional sign convention,
\begin{align}
    (R_{p(u)}q_u)^\top (R_{p(v)}k_v)
    &=
    q_u^\top R_{p(u)}^\top R_{p(v)}k_v \\
    &=
    q_u^\top R_{p(v)-p(u)}k_v.
\end{align}
Thus row-major RoPE is a bilinear relative-score instance with
\[
    M_d=R_d
\]
up to sign convention and the usual attention scaling constant, which can be
absorbed into \(M_d\). Therefore, if the RoPE relative matrices distinguish two
offsets in the same \(D_4\)-offset class, such as \(1\) and \(9\) on a
\(9\times9\) grid, the RoPE score does not satisfy the tying condition required
by Theorem~\ref{thm:rigidity}. Hence such a nondegenerate
row-major RoPE head does not by itself certify exact \(D_4\)-equivariance.
\end{proof}

\paragraph{Scope of the result.}
Theorem~\ref{thm:rigidity} is a component-level certificate
theorem. It assumes a single full, unmasked attention head, a row-major
relative-score form, and a position-only \(D_4\) action on hidden states. It
does not imply that a full multi-layer Transformer with MLPs, residual streams,
normalization, multi-head interactions, or learned internal representations
cannot learn a \(D_4\)-equivariant function from data. It says that standard
row-major relative-score attention does not enforce the \(D_4\) certificate by
itself unless its offset dependence collapses to the offset classes above.
If hidden channels transform under a nontrivial \(D_4\) representation, or if a
mask removes the self-edge \(v=u\), the necessary tying condition changes.

\subsection{Representation-wise denoising accounting}
\label{app:denoising-spectrum}

This appendix gives a denoising-accounting identity. The result is a
posterior \(L^2\) identity, not a KL identity. It tracks the nontrivial representation modes
of the hidden view label through the reverse denoising chain.

Let $V\sim\mathrm{Unif}(G)$.  For any random variable $S$, define the posterior
likelihood ratio of the view label by
\begin{equation}
    \alpha_S(g):=\frac{\Pr(V=g\mid S)}{\Pr(V=g)}=m\Pr(V=g\mid S).
\end{equation}
Then $m^{-1}\sum_g\alpha_S(g)=1$.  For $\rho\in\widehat G$, define
\begin{equation}
    \widehat\alpha_{S,\rho}
    :=\frac1m\sum_{g\in G}\alpha_S(g)\rho(g)^*.
\end{equation}
For nontrivial $\rho\ne\mathbf 1$, define the $\rho$-view energy in $S$ by
\begin{equation}
    \mathcal E_\rho(S)
    :=d_\rho\,\mathbb E_S
    \left[\|\widehat\alpha_{S,\rho}\|_{\mathrm{HS}}^2\right].
\end{equation}
For two random variables $S,T$, define the conditional innovation
\begin{equation}
    \mathcal J_\rho(T;V\mid S)
    :=d_\rho\,\mathbb E_{S,T}
    \left[
    \|\widehat\alpha_{S,T,\rho}-\widehat\alpha_{S,\rho}\|_{\mathrm{HS}}^2
    \right].
\end{equation}

\begin{lemma}[Posterior Pythagorean identity]
\label{lem:posterior-pythagorean}
For any $S,T$ and any nontrivial $\rho$,
\begin{equation}
    \mathcal E_\rho(S,T)
    =\mathcal E_\rho(S)+\mathcal J_\rho(T;V\mid S).
\end{equation}
The same identity holds with $S$ and $T$ interchanged.
\end{lemma}

\begin{proof}
Bayes posterior averaging gives
\begin{equation}
    \alpha_S(g)=\mathbb E[\alpha_{S,T}(g)\mid S].
\end{equation}
Fourier transform in the group variable is linear, so
\begin{equation}
    \widehat\alpha_{S,\rho}
    =\mathbb E[\widehat\alpha_{S,T,\rho}\mid S].
\end{equation}
The Hilbert-space conditional variance identity gives
\begin{equation}
\mathbb E\|A\|_{\mathrm{HS}}^2
=\mathbb E\|\mathbb E[A\mid S]\|_{\mathrm{HS}}^2
+\mathbb E\|A-\mathbb E[A\mid S]\|_{\mathrm{HS}}^2,
\end{equation}
with $A=\widehat\alpha_{S,T,\rho}$.  Multiplying by $d_\rho$ proves the result.
\end{proof}

Now run the reverse denoising chain under input $Vx$ and pull each state back to
canonical coordinates:
\begin{equation}
    Z_H^c,Z_{H-1}^c,\ldots,Z_0^c,Y^c.
\end{equation}
For each step define
\begin{equation}
    A_h^\rho:=\mathcal J_\rho(Z_{h-1}^c;V\mid Z_h^c),
    \qquad
    B_h^\rho:=\mathcal J_\rho(Z_h^c;V\mid Z_{h-1}^c),
\end{equation}
and for the readout define
\begin{equation}
    A_R^\rho:=\mathcal J_\rho(Y^c;V\mid Z_0^c),
    \qquad
    B_R^\rho:=\mathcal J_\rho(Z_0^c;V\mid Y^c).
\end{equation}

\begin{theorem}[Representation-wise denoising accounting]
\label{thm:appendix-denoising-spectrum}
For every nontrivial representation $\rho\ne\mathbf 1$,
\begin{equation}
    \mathcal E_\rho(Y^c)
    =\mathcal E_\rho(Z_H^c)
    +\sum_{h=1}^H(A_h^\rho-B_h^\rho)
    +A_R^\rho-B_R^\rho.
    \label{eq:appendix-denoising-exact}
\end{equation}
Consequently,
\begin{equation}
    \mathcal E_\rho(Y^c)
    \le
    \mathcal E_\rho(Z_H^c)+\sum_{h=1}^H A_h^\rho+A_R^\rho.
    \label{eq:appendix-denoising-upper}
\end{equation}
\end{theorem}

\begin{proof}
Apply Lemma~\ref{lem:posterior-pythagorean} to
$(S,T)=(Z_h^c,Z_{h-1}^c)$ and to $(S,T)=(Z_{h-1}^c,Z_h^c)$.  The two expressions
for $\mathcal E_\rho(Z_h^c,Z_{h-1}^c)$ give
\begin{equation}
    \mathcal E_\rho(Z_{h-1}^c)-\mathcal E_\rho(Z_h^c)
    =A_h^\rho-B_h^\rho.
\end{equation}
Summing over $h=1,\ldots,H$ gives
\begin{equation}
    \mathcal E_\rho(Z_0^c)
    =\mathcal E_\rho(Z_H^c)
    +\sum_{h=1}^H(A_h^\rho-B_h^\rho).
\end{equation}
Apply the same argument to $(Z_0^c,Y^c)$ to obtain
\begin{equation}
    \mathcal E_\rho(Y^c)-\mathcal E_\rho(Z_0^c)=A_R^\rho-B_R^\rho.
\end{equation}
Combining the two displays proves Eq.~(\ref{eq:appendix-denoising-exact}).
Since all $B_h^\rho$ and $B_R^\rho$ are nonnegative by definition, dropping them
proves Eq.~(\ref{eq:appendix-denoising-upper}).
\end{proof}

\begin{corollary}[A nonzero output mode requires a nonzero source]
\label{cor:nonzero-source}
Assume the initial canonical noise state carries no $\rho$-view mode,
$\mathcal E_\rho(Z_H^c)=0$, and the readout injects no new $\rho$-view mode,
$A_R^\rho=0$.  If $\mathcal E_\rho(Y^c)>0$, then at least one reverse denoising
step has $A_h^\rho>0$.  In particular,
\begin{equation}
    \max_{1\le h\le H} A_h^\rho
    \ge \frac{\mathcal E_\rho(Y^c)}{H}.
\end{equation}
\end{corollary}

\begin{proof}
Under the assumptions, Eq.~(\ref{eq:appendix-denoising-upper}) becomes
\begin{equation}
    \mathcal E_\rho(Y^c)\le \sum_{h=1}^H A_h^\rho.
\end{equation}
If the left side is positive, the sum is positive, so at least one term is
positive.  The displayed lower bound follows by averaging.
\end{proof}

If the canonicalized reverse transition at step $h$ is independent of $V$ given
$Z_h^c$, then $Z_{h-1}^c\perp V\mid Z_h^c$, which implies $A_h^\rho=0$ for every
nontrivial $\rho$.  Thus $A_h^\rho>0$ is a one-way witness: the transition cannot
be view-independent on the posterior support.  The converse need not hold,
because a transition may depend on the view in a way that is outside the
posterior support or invisible to the particular representation mode.

\subsection{Finite-sample measurement and calibration}
\label{app:finite-k}

This appendix explains why the stochastic metrics are averaged over base states
and why pseudo-orbits are reported.  With $K$ samples per view, empirical maxima
and minima are biased: maxima tend to be too high and minima tend to be too low.
This can create a gap even when all views have the same true success
probability.

For fixed $x$, let $\widehat p_{g,K}(x)$ be the empirical mean of $K$ Bernoulli
success samples with mean $p_g(x)$.  Hoeffding's inequality and a union bound give
\begin{equation}
    \Pr\left[
    \max_{g\in G}|\widehat p_{g,K}(x)-p_g(x)|>\epsilon
    \right]
    \le 2|G|e^{-2K\epsilon^2}.
\end{equation}
On the complementary event,
\begin{equation}
    |\widehat C_{\max,K}(x)-C_{\max}(x)|\le\epsilon,
    \qquad
    |\widehat R_{\min,K}(x)-R_{\min}(x)|\le\epsilon.
\end{equation}
This controls concentration, but it does not remove extrema bias.

The canonical pseudo-orbit supplies a same-input reference for finite-sample extrema.  For each base state
$x$, draw $|G|K$ samples from the canonical layout only.  Randomly split them
into $|G|$ groups of size $K$ and compute the same max/min metrics as if those
groups were views.  Because all pseudo-views come from the same layout, any
pseudo gap is due to finite-sample extrema.  The real-minus-pseudo contrast compares the observed orbit against that same-input reference; it is not an unbiased correction for arbitrary heterogeneous view success rates.

Paired bootstrap intervals keep each base puzzle and its views together. The reported resampling units and training repetitions are specified with each study.

\subsubsection{Unbiased success dispersion}
\label{app:unbiased-proof}
Fix a trained model and a base problem $x$. Let $S^{(k)}\in\{0,1\}^m$ be the correctness vector of replicate $k$, with common expectation $p$. Replicate vectors are independent; their components need not be independent. Invalid outputs contribute zeros. Set $H=I_m-\mathbf1\mathbf1^\top/m$.
\begin{proposition}[Unbiasedness at every finite replication budget]
For $K\ge2$, the estimator in Eq.~\ref{eq:unbiased-v} satisfies
\[
\E[\widehat V(x)\mid x,\theta]=\frac1m p^\top Hp=V(x).
\]
\end{proposition}
\begin{proof}
For $k\ne\ell$, independence gives
\[
\E[S^{(k)\top}HS^{(\ell)}]=\sum_{g,h}H_{gh}\E[S_g^{(k)}]\E[S_h^{(\ell)}]=p^\top Hp=mV(x).
\]
There are $K(K-1)$ ordered distinct pairs. Dividing their sum by $mK(K-1)$ proves the result.
\end{proof}
An equivalent computation sets $c^{(k)}=HS^{(k)}$ and uses
\[
\widehat V=\frac{\|\sum_k c^{(k)}\|^2-\sum_k\|c^{(k)}\|^2}{mK(K-1)}.
\]
Negative estimates are retained. The $K(K-1)$ products are dependent and are not additional independent data. By linearity the puzzle-averaged estimate of $J=\E_x[V_C-V_D-V_A+V_B]$ is unbiased conditional on fixed trained models. A positive $J$ is an empirical result, not a conclusion of the proposition. Unbiasedness concerns expectation; finite-$K$ variance remains, and it does not imply that a finite realized $K'$ curve is flat or that an extrema statistic has converged.

This is an application of the standard independent-product/U-statistic construction used in distribution comparisons~\citep{gretton2012kernel}, not a new general principle of unbiased estimation. For a finite-second-moment feature map $\phi$ on canonicalized outputs, the same argument estimates
\[
D_\phi(x)=\frac1m\sum_g\|\E\phi(Y_g)-m^{-1}\sum_h\E\phi(Y_h)\|^2
\]
by cross-replicate products of view-centered features. Whole-output one-hot features identify full discrete-law equality; token one-hot features identify only token marginals. No full-output feature experiment is claimed here. In particular $V=0$ alone does not establish equality of output laws.

\subsubsection{An exact competence-conditioned null}
\label{app:conditional-null}
The following result needs the stronger condition that all $mK$ Bernoulli decodes for an arm and problem are mutually independent. Under $H_0:p_g(x)=p(x)$ for every view, let $n_g$ be the number of successes in view $g$ and $t=\sum_g n_g$.
\begin{proposition}[Conditional sampling law]
For $0\le n_g\le K$ and $\sum_g n_g=t$,
\begin{equation}
\Pr((n_g)_g\mid t,H_0)=\frac{\prod_g\binom K{n_g}}{\binom{mK}{t}}.
\label{eq:conditional-null}
\end{equation}
\end{proposition}
\begin{proof}
The unconditional mass of $(n_g)$ is $\prod_g\binom K{n_g}p^t(1-p)^{mK-t}$. The total count is binomial with mass $\binom{mK}{t}p^t(1-p)^{mK-t}$. Their ratio gives Eq.~\ref{eq:conditional-null}. Totals $t=0,mK$ are point masses.
\end{proof}
The unknown common success probability cancels separately for each problem and trained model. Resampling these counts and recomputing a statistic supplies an arm-wise equal-view null. Randomization $p$-values use the plus-one convention; a finite simulation has finite resolution. An arm-wise rejection is not a test of a favorable factorial interaction. Cross-arm null simulations must preserve any cross-arm decoding dependence. The weaker independent-replicate condition sufficient for $\widehat V$ does not automatically validate this conditional law.

A separate pooled-success plug-in calculation simulates independent $\operatorname{Bin}(K,\widehat p)$ counts, with $\widehat p=t/(mK)$, and computes an extrema reference. It is not the conditional hypergeometric law and is not an unbiased correction under unequal view success probabilities. We therefore do not interpret its attenuation as an exact fraction of an effect caused by sampling.

\subsubsection{Conditional calibration of token-marginal leakage}
\label{app:token-null}
For each puzzle, arm, and training seed, canonicalize the $mK$ complete outputs first. Pool them and uniformly repartition the individual whole outputs into $m$ groups of $K$. Recompute exactly the same smoothing and token-marginal KL as for the observed view partition. Under equal canonicalized output laws and exchangeability of decodes, conditioning on the pooled outputs makes these assignments an appropriate permutation reference. Moving whole outputs preserves all within-grid token dependence. Merely permuting the $m$ view bins leaves the symmetric KL statistic unchanged and cannot provide this reference.

We report observed leakage minus the Monte Carlo mean of the repartition statistic. Under alternatives this quantity is a conditional-reference contrast, not an unbiased estimator of population KL. Equality of token marginals is weaker than equality of the whole output law. Sampling and puzzle uncertainty are reported separately in Appendix~\ref{app:measurement-results}.

\subsection{One-call uniform symmetrization}
\label{app:symmetrization}
Let a finite group $G$ act on inputs and outputs, with transported correct target $y^\star(gx)=g y^\star(x)$. Draw $h$ uniformly, decode $Y\sim K_\theta(\cdot\mid hx)$, and return $h^{-1}Y$. For an output event $B$, the resulting law is
\[
K_{\mathrm{sym}}(B\mid x)=\frac1{|G|}\sum_h K_\theta(hB\mid hx).
\]
\begin{proposition}[Distributional equivariance with unchanged orbit mean]
$K_{\mathrm{sym}}(gB\mid gx)=K_{\mathrm{sym}}(B\mid x)$ for every $g$, and its success probability in every view equals $\bar p(x)$ of the original decoder.
\end{proposition}
\begin{proof}
In $|G|^{-1}\sum_h K_\theta(hgB\mid hgx)$, reindex $u=hg$ to obtain the original sum. Apply the identity to the transported correct target; the uniform mixture averages its original per-view success probabilities.
\end{proof}
Thus uniform random-view decoding preserves orbit-mean solving but can improve population worst-view success. Independent randomized queries need not agree pathwise. For correctness,
\[
\bar p(1-\bar p)=m^{-1}\sum_g p_g(1-p_g)+V(x),
\]
by expansion. Symmetrization can transfer between-view variation into within-query randomness without adding solving competence. This result does not constrain nonuniform selection or voting schemes.

\section{Additional Results}
\label{app:additional}

\subsection{Metric definitions and controlled scaling}
\label{app:metric-summary}\label{app:scaling}
\begin{figure}[t]
\centering
\begin{minipage}[t]{0.50\linewidth}
  \centering
  \vspace{0pt}
  \includegraphics[width=\linewidth]{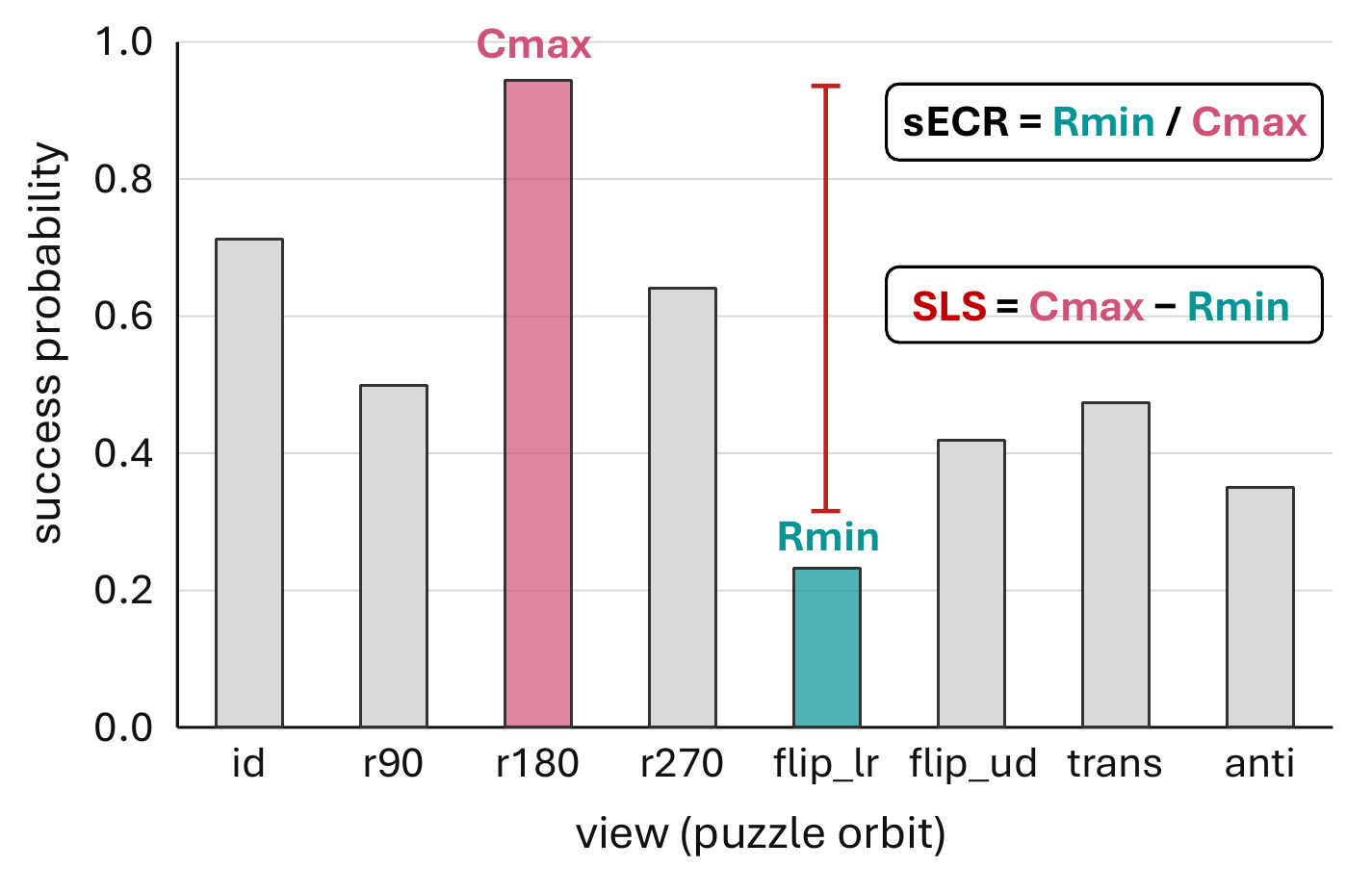}
  \caption{\small \textbf{Visualizing stochastic orbit metrics.}
  For each logical state, we estimate success under every $D_4$ view after canonicalizing predictions back to the original board.
  $C_{\max}$ captures best-view capacity, $R_{\min}$ captures worst-view reliability, and sECR measures the fraction of best-view capacity retained in the worst view.}
  \label{fig:stochastic_metrics}
\end{minipage}\hfill
\begin{minipage}[t]{0.46\linewidth}
  \centering
  \vspace{0pt}
  \captionof{table}{\small \textbf{Metrics used in this paper.}
  Recorded metrics use one prediction per view. Sampled metrics use $K$ samples per view, and SLS includes finite-sample extrema bias.}
  \label{tab:orbit-metrics-summary}
  \footnotesize
  \setlength{\tabcolsep}{3pt}
  \begin{tabular}{llc}
  \toprule
  Metric & Definition & \\
  \midrule
  \multicolumn{3}{l}{\emph{Recorded (Table~\ref{tab:large-dllm})}}\\
  Acc & mean correctness over views & $\uparrow$\\
  ANY & correct in at least one view & \\
  ALL & correct in every view & $\uparrow$\\
  cLE & $100\,\mathrm{ALL}/\mathrm{ANY}$ & $\uparrow$\\
  \midrule
  \multicolumn{3}{l}{\emph{Sampled (Tables~\ref{tab:controlled-scaling-ac}, \ref{tab:combined_post_training})}}\\
  MeanSucc & mean success over views & $\uparrow$\\
  $\widehat C_{\max}$ & estimated best-view success & $\uparrow$\\
  $\widehat R_{\min}$ & estimated worst-view success & $\uparrow$\\
  SLS & $\widehat C_{\max}-\widehat R_{\min}$ & $\downarrow$\\
  sECR & $\widehat R_{\min}/\widehat C_{\max}$ & $\uparrow$\\
  $\widehat L_{\mathrm{tok}}$ & token-marginal view leakage & $\downarrow$\\
  $\widehat V$ & success-probability dispersion & $\downarrow$\\
  \bottomrule
  \end{tabular}
\end{minipage}
\end{figure}

The metric overview appears in Figure~\ref{fig:stochastic_metrics} and Table~\ref{tab:orbit-metrics-summary}. Original pretrained scaling evaluations are reported separately from post-training repetitions. From 5M to 50M, mean AC24 success increases from 12.76\% to 32.85\%; the 100M checkpoint does not improve further. Sampled extrema include finite-$K$ bias and do not on their own identify population dispersion.

\begin{table}[t]
\centering\small
\caption{\textbf{Controlled MDLM scaling.} Pretrained models evaluated with $K=16$ samples per view. Success rates, sECR, and SLS are percentages; leakage is in nats. SLS $=\widehat C_{\max}-\widehat R_{\min}$ includes finite-sample extrema bias.}
\label{tab:controlled-scaling-ac}
\setlength{\tabcolsep}{4pt}
\begin{tabular}{llrrrrr}
\toprule
Params & Window & MeanSucc $\uparrow$ & $\widehat R_{\min}$ $\uparrow$ & sECR $\uparrow$ & SLS $\downarrow$ & $\widehat L_{\rm tok}$ $\downarrow$\\
\midrule
5M   & AC16 & 16.87 &  6.18 & 20.53 & 23.91 & 0.096\\
     & AC24 & 12.76 &  3.69 & 14.89 & 21.07 & 0.097\\
\midrule
10M  & AC16 & 30.13 & 16.20 & 35.52 & 29.41 & 0.083\\
     & AC24 & 26.31 & 12.88 & 31.08 & 28.56 & 0.085\\
\midrule
50M  & AC16 & \textbf{36.27} & \textbf{21.61} & \textbf{41.24} & 30.79 & \textbf{0.076}\\
     & AC24 & \textbf{32.85} & \textbf{18.71} & \textbf{38.53} & 29.86 & \textbf{0.077}\\
\midrule
100M & AC16 & 34.61 & 20.32 & 40.61 & 29.72 & \textbf{0.076}\\
     & AC24 & 31.07 & 17.24 & 37.15 & 29.16 & \textbf{0.077}\\
\bottomrule
\end{tabular}
\end{table}

The original pretrained checkpoint's real-minus-pseudo spread is 1.24 points, compared with 11.11 after RSM and 4.05 after RSM+PR (Table~\ref{tab:initial-full}). This is a same-input reference diagnostic, not an unbiased decomposition. It does not summarize all post-training behavior.

\subsection{Per-transformation breakdown for code and mathematics}
\label{app:per-transform}

\begin{table}[htbp]
\centering
\caption{\small \textbf{Per-transformation \textsc{Any}/\textsc{All}} (percentages) for the paired evaluations in Table~\ref{tab:large-dllm}.}
\label{tab:per-transform}
\vskip 0.1in
\resizebox{\linewidth}{!}{
\begin{tabular}{l l cccc}
\toprule
Task & Transformation & LLaDA-8B & Dream-7B & iLLaDA-8B & LLaDA-MoE \\
\midrule
Code infilling & Identifier renaming & 59.4 / 34.1 & 65.9 / 34.9 & 43.1 / 17.6 & 62.8 / 32.4 \\
& Operand exchange & 55.7 / 34.3 & 61.3 / 37.7 & 41.6 / 20.0 & 58.9 / 32.6 \\
& Statement swap & 54.2 / 38.2 & 56.9 / 33.8 & 40.0 / 23.1 & 54.2 / 28.9 \\
& Statement relocation & 54.1 / 30.7 & 63.3 / 32.8 & 46.1 / 19.6 & 60.3 / 26.7 \\
& Inert-line insertion & 58.3 / 24.0 & 63.5 / 24.0 & 45.7 / 12.2 & 61.6 / 20.5 \\
\midrule
MATH-500 & Variable renaming & 38.3 / 20.4 & 22.6 / 11.5 & 34.7 / 19.6 & 45.1 / 23.0 \\
& Premise permutation & 31.8 / 11.4 & 25.0 / 9.1 & 25.6 / 14.0 & 45.5 / 18.2 \\
& Identity-operation insertion & 39.9 / 19.8 & 26.7 / 12.5 & 35.6 / 17.6 & 49.5 / 20.5 \\
\bottomrule
\end{tabular}
}
\end{table}

For LLaDA-8B under variable renaming, accuracy rises from 28.5\% to 30.2\% while 8.1\% of all instances change from correct to wrong and 9.8\% from wrong to correct; the corresponding \textsc{Any}$-$\textsc{All} of 17.9 points equals $r+b$ in Eq.~(\ref{eq:two-view-identity}).

\subsection{Same-input decoding controls across tasks}
\label{app:decoding-controls}
These are new evaluations, separate from the original greedy summaries in Table~\ref{tab:large-dllm}. For each base problem a pseudo-orbit uses $m$ independent calls to its original presentation, whereas a real orbit uses one call to each of $m$ presentations ($m=8$ for Sudoku and $m=2$ for code/MATH). Calls and decoding settings agree within each comparison. Under equal success probability across views and independent decoding, the joint correctness law, and hence the ANY/ALL distribution, is the same. A real-minus-pseudo gap is not an unbiased additive decomposition of noise under arbitrary heterogeneous view distributions. Intervals resample base problems (10,000 resamples), keeping all views and pairs of a problem together. The 95\% intervals exclude zero for every Sudoku and code row and for MATH LLaDA-8B and Dream-7B. MATH uses temperature 0.2 across models because Dream produced empty strings in the temperature-0.5 control.
\begin{table}[htbp]
\centering\small
\caption{\textbf{Full large-model decoding control.} Values are ALL/ANY in percent. Gap increases are computed from the unrounded analysis, so may differ by 0.1 from subtraction of displayed entries. Sudoku/code use temperature 0.5; MATH uses 0.2. $\dagger$ denotes 200 of 582 admitted pairs.}
\label{tab:decoding-controls}
\setlength{\tabcolsep}{4pt}
\begin{tabular}{llccc}
\toprule
Task & Model & Same input & Different views & Gap increase (pp)\\
\midrule
Sudoku AC16 & LLaDA-8B & 36.2/52.0 & 1.3/98.6 & +81.5\\
Sudoku AC24 & LLaDA-8B & 11.4/22.4 & 0.0/73.5 & +62.5\\
Sudoku AC32 & LLaDA-8B & 1.4/6.2 & 0.0/26.6 & +21.8\\
Sudoku AC24 & Dream-7B & 75.0/89.8 & 50.5/100.0 & +34.7\\
Sudoku AC32 & Dream-7B & 55.1/74.4 & 17.4/99.9 & +63.2\\
Sudoku AC32 & iLLaDA-8B & 71.8/75.4 & 23.0/99.0 & +72.4\\
Sudoku AC32 & LLaDA-MoE & 48.6/68.6 & 6.3/99.8 & +73.5\\
\midrule
Code & LLaDA-8B & 37.3/53.6 & 28.5/57.6 & +12.8\\
Code & Dream-7B & 40.9/59.3 & 29.3/62.7 & +15.0\\
Code & iLLaDA-8B & 21.7/36.0 & 15.3/43.2 & +13.6\\
Code & LLaDA-MoE & 33.6/59.0 & 25.8/60.7 & +9.5\\
\midrule
MATH & LLaDA-8B & 16.8/29.4 & 16.2/33.7 & +5.0\\
MATH & Dream-7B & 14.3/17.5 & 9.1/19.2 & +6.8\\
MATH & iLLaDA-8B$^\dagger$ & 17.3/30.5 & 14.2/30.5 & +3.1\\
MATH & LLaDA-MoE & 18.7/39.5 & 17.5/42.1 & +3.8\\
\bottomrule
\end{tabular}
\end{table}

\subsection{Autoregressive comparison}
\label{app:ar-ladder}
We evaluate Qwen2.5-7B-Instruct and Llama-3.1-8B-Instruct greedily. Each completed Sudoku setting contains 500 problems and eight D4 views; parsing failures remain in the denominator. Different anchored windows are not used to rank model families.
\begin{table}[htbp]
\centering\small
\caption{\textbf{AR Sudoku controls.} Percentages over all recorded outputs or complete orbits, as indicated. ``NT'' means parse failures were not separately tallied, not zero.}
\label{tab:ar-sudoku}
\begin{tabular}{llrrrr}
\toprule
Model & Window & Mean success & ANY & ALL & Parse failures\\
\midrule
Qwen2.5-7B & AC4 & 57.525 & 100.0 & 1.6 & 663/4,000\\
Qwen2.5-7B & AC16 & 6.700 & 42.6 & 0.0 & NT\\
Llama-3.1-8B & AC2 & 27.375 & 87.0 & 0.0 & 25/4,000\\
Llama-3.1-8B & AC4 & 13.2 & 63.8 & 0.0 & NT\\
\bottomrule
\end{tabular}
\end{table}
Qwen AC4 has 2,301 correct and 1,036 parsed-incorrect outputs. Every puzzle has a correct view, so it has at most seven parsed-incorrect views. At least $\lceil1036/7\rceil=148$ puzzles therefore contain both a parsed-correct and a parsed-incorrect view. For Llama AC2, 435 puzzles have mixed correctness and at most 25 puzzles can contain a parsing failure; at least 410/500 have all eight outputs parsed and still mix correctness. Parsing may affect the aggregate, but cannot explain all presentation-dependent failures.

\paragraph{MATH aggregation.}
The 328 base problems supply 582 admitted canonical--transformed pairs and 910 unique input texts. There are 105 problems with two presentations, 192 with three, and 31 with four. Every text has one greedy prediction. For Table~\ref{tab:ar-math}, we first compute each problem's mean correctness, ANY, and ALL over all its available presentations and then average each quantity equally over 328 problems. Accuracy is not a 910-input micro-average and need not equal $(\mathrm{ANY}+\mathrm{ALL})/2$ when orbit sizes exceed two.
\begin{table}[htbp]
\centering\small
\caption{\textbf{AR MATH, base-problem-balanced available-presentation summaries.} Percentages. The gap is the fraction of base problems with mixed correctness.}
\label{tab:ar-math}
\begin{tabular}{lrrrr}
\toprule
Model & Accuracy & ANY & ALL & ANY$-$ALL\\
\midrule
Qwen2.5-7B & 38.1 & 47.0 & 29.9 & 17.1\\
Llama-3.1-8B & 32.3 & 43.9 & 22.6 & 21.3\\
\bottomrule
\end{tabular}
\end{table}
Qwen has 25/910 parsing failures and Llama 10/910, all scored incorrect. Their input-micro accuracies are 336/910 (36.9231\%) and 287/910 (31.5385\%). For comparison using the distinct 582-pair weighting, Accuracy/ANY/ALL are 36.00/41.58/30.41 for Qwen and 31.70/39.00/24.40 for Llama. The pair-count numerators are $(419/1164,242/582,177/582)$ and $(369/1164,227/582,142/582)$; only this pair weighting obeys the two-view identity. The LLaDA pair summary is a separate aggregation and includes 112/582 discordant pairs. We do not compare orbit-level and pair-level magnitudes as if they were identically weighted.

These controls extend the diagnostic's scope, consistent with AR consistency research~\citep{lai2026multidimensional}. The masked-diffusion result shows that causal masking is not necessary for presentation dependence. Generation order within one serialization and transport across equivalent serializations are distinct. The full, unmasked-attention assumptions of Theorem~\ref{thm:rigidity} are not extended to causal AR attention.

\subsection{Equal-budget inference-time baselines}
\label{app:inference}
Each inference procedure is applied to every original Table~\ref{tab:initial-full} recipe checkpoint with the same query budget. This study is distinct from the common-harness three-seed repetition. Calls below are per evaluated presentation; joint success requires correct outcomes in all eight presentations of a problem. The eight samples used to solve a query are not the $K=16$ budget used to estimate symbolic success probabilities.
\begin{table}[htbp]
\centering\small
\caption{\textbf{Inference controls on fixed post-trained checkpoints.} Mean success / joint success, in percent. Voting is cell-wise, not whole-board voting; verify-and-select uses the same task checker in every arm.}
\label{tab:inference}
\setlength{\tabcolsep}{4pt}
\begin{tabular}{lcccc}
\toprule
Recipe & Native (1) & Random D4 (1) & Cell vote (8) & Verify-select (8)\\
\midrule
Canonical & 42.9/7.0 & 42.1/5.8 & 39.5/18.4 & 91.6/68.2\\
PR & 42.3/6.6 & 41.4/6.2 & 37.7/16.2 & 91.1/65.2\\
RSM & 46.3/7.4 & 44.5/7.8 & 42.5/16.6 & 91.7/73.8\\
RSM+PR & 46.8/7.6 & 46.3/8.4 & 44.3/19.2 & 92.0/72.6\\
\bottomrule
\end{tabular}
\end{table}
For verify-and-select, RSM+PR minus Canonical joint success is $+4.40\ [1.20,7.60]$ points and minus PR is $+7.40\ [4.00,11.00]$. These puzzle-paired intervals condition on the checkpoints. RSM alone has the largest joint point estimate. At native one-call decoding, the mean-success contrasts are $+3.93\ [2.68,5.18]$ and $+4.48\ [3.18,5.78]$; these are this inference study's estimates, not the $K=16$ recipe contrasts. Cell-wise voting lowers mean success for every recipe. Uniform D4 symmetrization preserves the population orbit mean (Appendix~\ref{app:symmetrization}); finite sampled means need not be identical. This experiment establishes recipe differences under equal inference procedures, not that training is necessary for symmetrization or cheaper at every deployment volume.

\subsection{Original recipe diagnostics and new seed repetitions}
\label{app:posttrain-ci}

\begin{table}[htbp]
\centering\small
\caption{\textbf{Complete initial 50M recipe diagnostics.} Success and spread values are percentages; token leakage is in nats. Real-minus-pseudo contrasts use a canonical repeated-input reference.}
\label{tab:initial-full}
\resizebox{\linewidth}{!}{
\begin{tabular}{lrrrrrrrr}
\toprule
Recipe & MeanSucc & $\widehat C_{\max}$ & $\widehat R_{\min}$ & sECR & SLS & $\widehat L_{\rm tok}$ & RobustGap & SLSGap\\
\midrule
Pretrained & 31.53 & 47.29 & 17.41 & 36.82 & 29.88 & 0.0775 & $-0.51$ & 1.24\\
Canonical 80k & 41.58 & 57.79 & 25.94 & 44.88 & 31.85 & 0.0712 & $-1.69$ & 2.79\\
Canonical 10k & 43.12 & 59.76 & 27.43 & 45.89 & 32.34 & 0.0714 & $-1.56$ & 3.64\\
PR & 42.16 & 58.46 & 26.64 & 45.56 & 31.82 & 0.0709 & $-1.20$ & 2.88\\
PR grouped & 41.15 & 57.35 & 25.60 & 44.64 & 31.75 & 0.0709 & $-1.27$ & 2.16\\
RSM & 45.74 & 65.81 & 26.16 & 39.75 & 39.65 & 0.0783 & $-6.04$ & 11.11\\
RSM+PR & 47.00 & 63.89 & 30.50 & 47.74 & 33.39 & 0.0684 & $-2.34$ & 4.05\\
\bottomrule
\end{tabular}}
\end{table}

\begin{table}[htbp]
\centering
\caption{\small \textbf{Post-training endpoints with 95\% bootstrap intervals} (paired over base puzzles, 10,000 resamples). sECR and $\widehat R_{\min}$ are percentages; $\widehat L_{\mathrm{tok}}$ is in nats.}
\label{tab:posttrain-ci}
\vskip 0.1in
\resizebox{\linewidth}{!}{
\begin{tabular}{l ccc}
\toprule
Recipe & sECR$^{16}$ & $\widehat R_{\min}^{16}$ & $\widehat L_{\mathrm{tok}}$ \\
\midrule
Pretrained & 36.82 [34.44, 39.15] & 17.41 [15.79, 19.09] & 0.0775 [0.0755, 0.0794] \\
Canonical 80k & 44.88 [42.56, 47.14] & 25.94 [23.89, 28.01] & 0.0712 [0.0690, 0.0734] \\
Canonical 10k & 45.89 [43.58, 48.11] & 27.43 [25.34, 29.58] & 0.0714 [0.0691, 0.0737] \\
PR & 45.56 [43.27, 47.78] & 26.64 [24.56, 28.73] & 0.0709 [0.0686, 0.0732] \\
PR grouped & 44.64 [42.29, 46.88] & 25.60 [23.56, 27.66] & 0.0709 [0.0687, 0.0732] \\
RSM & 39.75 [37.31, 42.12] & 26.16 [24.05, 28.36] & 0.0783 [0.0755, 0.0811] \\
RSM+PR & \textbf{47.74} [45.44, 50.02] & \textbf{30.50} [28.30, 32.77] & \textbf{0.0684} [0.0660, 0.0709] \\
\bottomrule
\end{tabular}
}
\end{table}

\begin{table}[htbp]
\centering
\caption{\small \textbf{Original single-run paired contrasts} (percentage points unless noted; 95\% intervals). The interaction is $(\text{RSM+PR}-\text{RSM})-(\text{PR}-\text{Canonical})$.}
\label{tab:posttrain-contrasts}
\vskip 0.1in
\begin{tabular}{l l c}
\toprule
Contrast & Measure & Estimate \\
\midrule
RSM $-$ Canonical 10k & canonical-view success & $+2.0\ [+0.7, +3.3]$ \\
RSM $-$ Canonical 10k & sECR & $-6.1\ [-7.7, -4.6]$ \\
RSM $-$ Canonical 10k & $\widehat L_{\mathrm{tok}}$ (nats) & $+0.0069\ [+0.0054, +0.0084]$ \\
RSM+PR $-$ RSM & sECR & $+8.0\ [+6.4, +9.6]$ \\
PR $-$ Canonical 10k & sECR & $-0.3\ [-1.2, +0.6]$ \\
PR $-$ Canonical 80k & sECR & $+0.7\ [-0.1, +1.5]$ \\
Interaction (vs.\ Canonical 10k) & sECR & $+8.3\ [+6.6, +10.1]$ \\
Interaction (vs.\ Canonical 80k) & sECR & $+7.3\ [+5.6, +9.0]$ \\
RSM+PR $-$ Canonical 10k & sECR & $+1.9\ [+0.6, +3.1]$ \\
\bottomrule
\end{tabular}
\end{table}

Canonical-view success is MeanSucc$^{16}$ restricted to the identity view, the quantity a conventional single-layout protocol would report (43.82\% for canonical SFT and 45.81\% for RSM).

\subsubsection{Direct three-seed repetition of the four main recipes}
\label{app:recipe-seeds}
The following analysis uses only the twelve common batch-64 evaluations. There are 500 AC24 test puzzles, eight D4 views, and sixteen samples per view, with exact-match full-board scoring, temperature 1, 128 steps, and the \texttt{ddpm\_cache} decoder. Each recipe starts from the same pretrained 50M checkpoint and uses 9,000 updates. Seed 0 is the corresponding original checkpoint re-evaluated here; seeds 1 and 2 are additional fine-tunings. Original batch-512 sampled outputs are not mixed into this analysis. For example, original RSM+PR MeanSucc is 47.0047\%, whereas this seed-0 evaluation is 46.48\%.
\begin{table}[htbp]
\centering\small
\caption{\textbf{All twelve common-harness recipe evaluations.} All columns except sECR are percent; sECR is a ratio. SLS is computed from unrounded puzzle-level extrema and then averaged.}
\label{tab:recipe-all-seeds}
\begin{tabular}{lrrrrrr}
\toprule
Recipe & Seed & MeanSucc & $\widehat C_{\max}$ & $\widehat R_{\min}$ & SLS & sECR\\
\midrule
Canonical & 0 & 42.84 & 59.53 & 26.51 & 33.01 & .4454\\
 & 1 & 42.90 & 59.53 & 26.48 & 33.05 & .4448\\
 & 2 & 42.70 & 59.40 & 26.21 & 33.19 & .4413\\
PR & 0 & 41.76 & 57.71 & 26.03 & 31.69 & .4509\\
 & 1 & 41.54 & 57.68 & 25.81 & 31.86 & .4476\\
 & 2 & 41.74 & 57.78 & 25.98 & 31.80 & .4496\\
RSM & 0 & 45.60 & 65.58 & 25.66 & 39.91 & .3913\\
 & 1 & 45.52 & 65.63 & 25.46 & 40.16 & .3880\\
 & 2 & 45.56 & 65.60 & 25.56 & 40.04 & .3897\\
RSM+PR & 0 & 46.48 & 63.20 & 29.74 & 33.46 & .4705\\
 & 1 & 46.50 & 63.06 & 29.76 & 33.30 & .4720\\
 & 2 & 46.58 & 63.28 & 29.86 & 33.41 & .4719\\
\bottomrule
\end{tabular}
\end{table}
\begin{table}[htbp]
\centering\small
\caption{\textbf{Common-harness recipe contrasts.} 95\% intervals use 10,000 resamples. Paired resampling fixes the three trained seed blocks and resamples puzzles; crossed resampling resamples both seed blocks and puzzles, sharing indices across arms. Success differences are points; sECR differences are ratios.}
\label{tab:recipe-new-contrasts}
\begin{tabular}{llrrr}
\toprule
Contrast & Metric & Estimate & Paired interval & Crossed interval\\
\midrule
RSM$-$Canonical & Mean & +2.75 & [2.03,3.46] & [2.03,3.45]\\
 & $\widehat R_{\min}$ & $-0.84$ & [$-1.80$,0.09] & [$-1.81$,0.12]\\
 & sECR & $-0.0541$ & [$-0.0686$,$-0.0398$] & [$-0.0692$,$-0.0390$]\\
PR$-$Canonical & Mean & $-1.13$ & [$-1.36$,$-0.91$] & [$-1.45$,$-0.85$]\\
 & $\widehat R_{\min}$ & $-0.46$ & [$-0.86$,$-0.07$] & [$-0.98$,0.02]\\
 & sECR & +0.0055 & [$-0.0020$,0.0132] & [$-0.0037$,0.0144]\\
RSM+PR$-$RSM & Mean & +0.96 & [0.41,1.52] & [0.40,1.51]\\
 & $\widehat R_{\min}$ & +4.23 & [3.35,5.10] & [3.35,5.13]\\
 & sECR & +0.0818 & [0.0676,0.0958] & [0.0672,0.0969]\\
RSM+PR$-$Canonical & Mean & +3.70 & [3.20,4.20] & [3.19,4.22]\\
 & $\widehat R_{\min}$ & +3.39 & [2.65,4.10] & [2.61,4.16]\\
 & sECR & +0.0277 & [0.0155,0.0397] & [0.0149,0.0404]\\
RSM+PR$-$PR & Mean & +4.84 & [4.30,5.37] & [4.30,5.40]\\
 & $\widehat R_{\min}$ & +3.85 & [3.12,4.60] & [3.10,4.61]\\
 & sECR & +0.0221 & [0.0101,0.0342] & [0.0092,0.0349]\\
Interaction & Mean & +2.0896 & [1.55,2.64] & [1.52,2.69]\\
 & $\widehat R_{\min}$ & +4.6875 & [3.79,5.60] & [3.71,5.71]\\
 & sECR & +0.076263 & [0.0609,0.0914] & [0.0594,0.0932]\\
\bottomrule
\end{tabular}
\end{table}
RSM+PR is highest in mean success, estimated worst-view success, and sECR in all three seeds. Mean interactions are 1.95/2.34/1.98 points (SD 0.21); estimated worst-view interactions are 4.56/4.96/4.54 (SD 0.24); sECR interactions are .0736/.0812/.0740 (SD .0043). Within each seed, sECR is the ratio of puzzle-averaged extrema, and these three ratios are then averaged. Every bootstrap replicate uses the same definition. The pooled-numerator/denominator alternative differs by at most $2\times10^{-6}$ here but is not the reported estimand.

RSM alone increases mean solving while its estimated worst-view difference is unresolved. Adding PR improves both mean and estimated worst-view success. The absolute gap need not rank recipes in the same order as normalized retention: the RSM+PR$-$Canonical SLS contrast is $+0.31$ points, crossed interval $[-0.70,1.32]$, while RSM+PR$-$PR SLS is $+1.61\ [0.60,2.64]$. We retain these diagnostics rather than claiming improvement in every stability summary.
\begin{table}[htbp]
\centering\small
\caption{\textbf{Auxiliary recipe diagnostics.} Point estimates and crossed 95\% intervals in percentage points. These retain the non-primary best-view and absolute-gap comparisons.}
\label{tab:recipe-auxiliary}
\begin{tabular}{lcc}
\toprule
Contrast & $\widehat C_{\max}$ & SLS\\
\midrule
RSM$-$Canonical & $+6.12\ [5.04,7.18]$ & $+6.95\ [5.71,8.19]$\\
PR$-$Canonical & $-1.76\ [-2.25,-1.27]$ & $-1.30\ [-1.98,-0.63]$\\
RSM+PR$-$RSM & $-2.42\ [-3.35,-1.48]$ & $-6.65\ [-7.90,-5.40]$\\
RSM+PR$-$Canonical & $+3.70\ [2.86,4.52]$ & $+0.31\ [-0.70,1.32]$\\
RSM+PR$-$PR & $+5.46\ [4.64,6.30]$ & $+1.61\ [0.60,2.64]$\\
Interaction & $-0.66\ [-1.65,0.36]$ & $-5.35\ [-6.75,-3.93]$\\
\bottomrule
\end{tabular}
\end{table}

\subsection{Spectral localization}
\label{app:spectral}

Applying the decomposition of Theorem~\ref{thm:appendix-view-spectrum} to token marginals splits the view-dependent energy $\widehat\chi^2_{\mathrm{tok}}$ across the four nontrivial irreducible representations of $D_4$: three sign characters (sgn, axis, diag) and the two-dimensional standard representation $E$. The axis character separates $\{e, r^2, \sigma_h, \sigma_v\}$, which keep rows as rows, from $\{r, r^3, \sigma_d, \sigma_{d'}\}$, which exchange rows and columns. An isotropic reference across the seven nontrivial Fourier dimensions assigns the Plancherel shares: 14.3\% for each character and 57.1\% for $E$.

\begin{table}[htbp]
\centering
\caption{\small \textbf{Share of token-level view disagreement by $D_4$ mode} (percent of $\widehat\chi^2_{\mathrm{tok}}$; point estimates from the same token marginals and smoothing as Table~\ref{tab:combined_post_training}).}
\label{tab:spectral}
\vskip 0.1in
\begin{tabular}{l c cccc}
\toprule
Recipe & Total $\widehat\chi^2_{\mathrm{tok}}$ & sgn & axis & diag & $E$ \\
\midrule
Pretrained & 0.1703 & 13.9 & 15.3 & 14.0 & 56.8 \\
Canonical 80k & 0.1580 & 13.9 & 15.6 & 14.2 & 56.3 \\
Canonical 10k & 0.1585 & 13.7 & 16.2 & 13.9 & 56.3 \\
PR & 0.1576 & 13.8 & 15.7 & 14.0 & 56.6 \\
PR grouped & 0.1572 & 13.9 & 15.7 & 14.0 & 56.4 \\
RSM & 0.1709 & 12.6 & \textbf{19.8} & 12.6 & 54.9 \\
RSM+PR & 0.1510 & 13.4 & 17.5 & 13.8 & 55.3 \\
\midrule
Flat reference & --- & 14.3 & 14.3 & 14.3 & 57.1 \\
\bottomrule
\end{tabular}
\end{table}

From the rounded table entries, RSM's total excess $\widehat\chi^2_{\mathrm{tok}}$ over Canonical 10k is 0.0124, of which approximately 0.0082 lies in the axis mode. Thus about two-thirds of the increase is concentrated in this component. Adding PR reduces the axis share to 17.5\% and the total estimate to 0.1510. These are spectral point estimates; the flat allocation is a reference, not a fitted sampling-null confidence interval.

\subsection{Additional Sudoku symmetry groups}
\label{app:more-groups}

We retrain the post-training suite from the same 50M checkpoint under four further answer-preserving groups: digit relabeling ($G_{\mathrm{digit}}$, which leaves every cell in place), row permutations within each band ($G_{\mathrm{row}}$), column permutations within each stack ($G_{\mathrm{col}}$), and band permutations ($G_{\mathrm{band}}$). For the larger groups, we sample eight group elements, matching $|D_4|$; for $G_{\mathrm{band}}$ we use all six. PR and RSM+PR draw views from the corresponding group.

\begin{table}[htbp]
\centering
\caption{\small \textbf{Post-training under additional symmetry groups.} Each entry is MeanSucc$^{16}$ (\%) / sECR$^{16}$ (fraction).}
\label{tab:more-groups}
\vskip 0.1in
\begin{tabular}{l cccc}
\toprule
Recipe & $G_{\mathrm{digit}}$ & $G_{\mathrm{row}}$ & $G_{\mathrm{col}}$ & $G_{\mathrm{band}}$ \\
\midrule
Pretrained & 31.5 / .362 & 31.5 / .381 & 31.9 / .382 & 31.7 / .433 \\
Canonical 80k & 41.8 / .468 & 41.9 / .480 & 42.1 / .478 & 42.1 / .521 \\
PR & 42.0 / .468 & 43.1 / .491 & 43.3 / .490 & 43.2 / .534 \\
RSM & 46.1 / .411 & 46.0 / .463 & 45.6 / .488 & 46.1 / .512 \\
RSM+PR & \textbf{47.2} / \textbf{.468} & \textbf{47.2} / \textbf{.495} & \textbf{46.5} / \textbf{.505} & \textbf{46.4} / \textbf{.537} \\
\bottomrule
\end{tabular}
\end{table}

In every group, RSM+PR has the highest mean success and the highest or tied-highest sECR, and PR raises sECR more on top of RSM ($+.057$, $+.032$, $+.017$, $+.025$) than on top of canonical SFT ($+.000$, $+.011$, $+.012$, $+.013$). Digit relabeling leaves the serialization unchanged, so the comparison also covers changes in symbol identity without moving cells.

\subsection{Maze completion}
\label{app:maze}
\begin{table}[htbp]
\centering\small
\caption{\textbf{Recipe utility on harder maze completion (n15).} Point estimates in percent. Parentheses give RSM+PR minus random-mask+PR. Models are initialized from the listed pretraining steps; both recipes use 5,000 post-training steps.}
\label{tab:maze-main}
\setlength{\tabcolsep}{4pt}
\begin{tabular}{llr>{\columncolor{myteal!15}}rr>{\columncolor{myteal!15}}r}
\toprule
& & \multicolumn{2}{c}{MeanSucc} & \multicolumn{2}{c}{$\widehat R_{\min}$}\\
\cmidrule(lr){3-4}\cmidrule(lr){5-6}
Model & Step & Random+PR & RSM+PR & Random+PR & RSM+PR\\
\midrule
50M & 100k & 91.33 & 97.46 (+6.13) & 80.44 & 92.50 (+12.06)\\
 & 150k & 94.21 & 98.70 (+4.49) & 85.38 & 96.12 (+10.74)\\
 & 200k & 96.54 & 99.26 (+2.72) & 90.25 & 97.56 (+7.31)\\
\midrule
100M & 100k & 83.53 & 94.68 (+11.15) & 67.94 & 86.75 (+18.81)\\
 & 150k & 88.28 & 95.42 (+7.14) & 75.38 & 88.75 (+13.37)\\
 & 200k & 90.46 & 96.97 (+6.51) & 78.81 & 91.50 (+12.69)\\
\bottomrule
\end{tabular}
\end{table}

Mazes are completed on a grid in which walls and start/end markers remain visible and corridor cells are the targets. $A_0$ masks a random proportion of corridor cells in the canonical orientation; PR ($B_0$) trains on an eight-view-expanded dataset; RSM trains on cached intermediate states from the maze solver with dense CE on every still-hidden cell; RSM+PR adds one online random $D_4$ view per instance. Each recipe post-trains for 5,000 steps from pretraining checkpoints at the listed steps and is evaluated on 100 held-out mazes with eight views and $K=16$ samples per view. The decoder uses 128 steps and temperature 1.0. With $S_{x,g,k}$ indicating full exact success, sampled joint success is
\begin{equation}
\widehat J_{8,16}=\frac1{16N}\sum_{x=1}^N\sum_{k=1}^{16}\prod_{g\in D_4}S_{x,g,k}.
\label{eq:maze-joint}
\end{equation}
This repeats the experiment of drawing one answer per view; $K$ is an estimation budget, not pass@$K$. Acc$^1$ equals MeanSucc here. Under conditional independence across views, the expectation is $\mathbb E_x\prod_gp_g(x)$, which becomes $\mathbb E_xp(x)^8$ when all view success probabilities agree. Consequently, sampled joint success is distinct from deterministic all-view correctness.

\begin{table}[htbp]
\centering\small
\caption{\textbf{Maze completion: harder (\texttt{n15}).} All entries are percentage point estimates. $\widehat J_{8,16}$ is sampled joint success, not deterministic ALL.}
\label{tab:maze-n15}
\begin{tabular}{lllrrrr}
\toprule
Model & Step & Recipe & MeanSucc & $\widehat R_{\min}$ & sECR & $\widehat J_{8,16}$\\
\midrule
50M & 100k & Random ($A_0$)    & 91.33 & 81.00 & 83.99 & 71.69\\
    &      & Random+PR ($B_0$) & 91.33 & 80.44 & 82.98 & 69.44\\
    &      & RSM               & 96.97 & 91.94 & 92.52 & 88.12\\
    &      & RSM+PR            & 97.46 & 92.50 & 92.85 & 89.19\\
\addlinespace[3pt]
50M & 150k & Random ($A_0$)    & 94.26 & 85.19 & 86.38 & 78.81\\
    &      & Random+PR ($B_0$) & 94.21 & 85.38 & 86.95 & 78.94\\
    &      & RSM               & 98.20 & 94.19 & 94.31 & 91.81\\
    &      & RSM+PR            & 98.70 & 96.12 & 96.19 & 94.12\\
\addlinespace[3pt]
50M & 200k & Random ($A_0$)    & 96.63 & 90.81 & 91.90 & 87.12\\
    &      & Random+PR ($B_0$) & 96.54 & 90.25 & 90.87 & 85.06\\
    &      & RSM               & 98.97 & 95.38 & 95.38 & 93.81\\
    &      & RSM+PR            & 99.26 & 97.56 & 97.56 & 95.81\\
\addlinespace[3pt]
100M & 100k & Random ($A_0$)    & 81.81 & 66.06 & 71.52 & 46.00\\
     &      & Random+PR ($B_0$) & 83.53 & 67.94 & 72.56 & 50.56\\
     &      & RSM               & 94.38 & 85.25 & 86.44 & 79.88\\
     &      & RSM+PR            & 94.68 & 86.75 & 87.96 & 81.25\\
\addlinespace[3pt]
100M & 150k & Random ($A_0$)    & 87.75 & 75.62 & 78.93 & 60.31\\
     &      & Random+PR ($B_0$) & 88.28 & 75.38 & 78.72 & 60.88\\
     &      & RSM               & 95.65 & 89.62 & 90.76 & 84.00\\
     &      & RSM+PR            & 95.42 & 88.75 & 89.87 & 84.06\\
\addlinespace[3pt]
100M & 200k & Random ($A_0$)    & 89.75 & 76.81 & 79.44 & 64.50\\
     &      & Random+PR ($B_0$) & 90.46 & 78.81 & 81.67 & 68.38\\
     &      & RSM               & 96.77 & 90.81 & 91.44 & 87.00\\
     &      & RSM+PR            & 96.97 & 91.50 & 92.02 & 87.56\\
\bottomrule
\end{tabular}
\end{table}

\begin{table}[htbp]
\centering\small
\caption{\textbf{Maze completion: easier (\texttt{n10}).} All entries are percentage point estimates. $\widehat J_{8,16}$ is sampled joint success, not deterministic ALL.}
\label{tab:maze-n10}
\begin{tabular}{lllrrr}
\toprule
Model & Step & Recipe & MeanSucc & $\widehat R_{\min}$ & $\widehat J_{8,16}$\\
\midrule
50M & 30k & Random ($A_0$)    & 97.42 & 92.06 & 87.25\\
    &     & Random+PR ($B_0$) & 97.33 & 92.19 & 86.62\\
    &     & RSM               & 99.48 & 97.56 & 96.62\\
    &     & RSM+PR            & 99.70 & 98.50 & 97.75\\
\addlinespace[3pt]
50M & 100k & Random ($A_0$)    & 99.35 & 97.19 & 95.81\\
    &      & Random+PR ($B_0$) & 99.54 & 97.88 & 96.62\\
    &      & RSM               & 99.66 & 98.12 & 97.62\\
    &      & RSM+PR            & 99.88 & 99.31 & 99.00\\
\addlinespace[3pt]
50M & 150k & Random ($A_0$)    & 99.62 & 98.31 & 97.31\\
    &      & Random+PR ($B_0$) & 99.80 & 98.88 & 98.50\\
    &      & RSM               & 99.83 & 99.00 & 98.75\\
    &      & RSM+PR            & 99.92 & 99.44 & 99.38\\
\addlinespace[3pt]
100M & 100k & Random ($A_0$)    & 98.91 & 96.12 & 93.75\\
     &      & Random+PR ($B_0$) & 99.03 & 96.50 & 94.56\\
     &      & RSM               & 99.74 & 98.75 & 98.12\\
     &      & RSM+PR            & 99.73 & 98.44 & 97.94\\
\addlinespace[3pt]
100M & 150k & Random ($A_0$)    & 99.09 & 96.19 & 94.25\\
     &      & Random+PR ($B_0$) & 99.46 & 97.19 & 96.31\\
     &      & RSM               & 99.77 & 98.62 & 98.31\\
     &      & RSM+PR            & 99.87 & 99.25 & 98.94\\
\addlinespace[3pt]
100M & 200k & Random ($A_0$)    & 99.34 & 97.12 & 95.75\\
     &      & Random+PR ($B_0$) & 99.44 & 97.50 & 96.56\\
     &      & RSM               & 99.85 & 99.19 & 98.88\\
     &      & RSM+PR            & 99.95 & 99.69 & 99.56\\
\bottomrule
\end{tabular}
\end{table}

On \texttt{n15}, RSM+PR exceeds random-mask+PR by 2.72--11.15 points in MeanSucc and 7.31--18.81 points in estimated worst-view success. Much of the improvement is already present with RSM alone. Adding PR increases $\widehat R_{\min}$ in five of six settings and lowers it by 0.87 points at 100M/150k. On \texttt{n10}, RSM+PR exceeds random-mask+PR in both success measures at every initialization. The comparisons across checkpoints describe initialization dependence, not independent training-seed replications.

\subsubsection{Density- and target-weight-matched maze comparison}
\label{app:maze-matched}
This new comparison is separate from the 100-maze initialization sweep above. It evaluates 500 held-out n15 mazes with three independent fine-tuning seeds per arm, a common training budget and decoder, density- and marginal-target-weight-matched states, and the same online PR procedure in both arms. The selected initialization rule is held common; it is not six independently pretrained models.
\begin{table}[htbp]
\centering\small
\caption{\textbf{Matched maze endpoint differences, RSM+PR minus random+PR.} Crossed seed-by-maze 95\% intervals. Success and sECR differences are percentage points; $\widehat V$ is success-probability variance.}
\label{tab:maze-matched}
\begin{tabular}{lcc}
\toprule
Metric & 50M & 100M\\
\midrule
Mean success & $+1.16\ [0.82,1.53]$ & $+2.10\ [1.58,2.67]$\\
Estimated $R_{\min}$ & $+1.52\ [1.03,2.05]$ & $+2.43\ [1.70,3.15]$\\
sECR & -- & $+0.74\ [-0.17,1.63]$\\
$\widehat V$ & -- & $+0.00075\ [0.00021,0.00129]$\\
\bottomrule
\end{tabular}
\end{table}
The 100M mean endpoints are 92.0\% and 94.1\%; the 50M endpoints are approximately 96.0\% and 97.1\%. The 100M plug-in calibrated estimated-worst-view contrast is $-0.42\ [-0.94,0.06]$, unresolved. Success dispersion is larger, not smaller, under RSM+PR in this comparison. The supported conclusion is a matched solving advantage, not a normalized-stability or success-dispersion advantage. Missing cells are not imputed.

\subsection{8B recipe results}
\label{app:8b}

Table~\ref{tab:8b} reports the full LLaDA-8B results on Sudoku-AC48. Each branch fine-tunes LLaDA-8B-Instruct with LoRA for 8,000 steps, and the checkpoint is selected on validation accuracy. RSM raises all-view correctness from 47.7\% to 74.8\%, and adding PR raises it further to 86.1\%, while PR alone leaves it at 46.0\%.

\begin{table}[htbp]
\centering
\caption{\small \textbf{RSM+PR on LLaDA-8B-Instruct} (LoRA, Sudoku-AC48, percentages). Each branch trains for 8,000 steps; checkpoints are selected on validation.}
\label{tab:8b}
\vskip 0.1in
\begin{small}
\begin{tabular}{l ccc}
\toprule
Recipe & Acc.\ $\uparrow$ & \textsc{Any} $\uparrow$ & \textsc{All} $\uparrow$ \\
\midrule
Canonical SFT & 88.2 & 99.8 & 47.7 \\
PR & 87.2 & \textbf{100.0} & 46.0 \\
RSM & 94.0 & \textbf{100.0} & 74.8 \\
\rowcolor{myteal!15} RSM+PR & \textbf{96.6} & \textbf{100.0} & \textbf{86.1} \\
\bottomrule
\end{tabular}
\end{small}
\vskip -0.1in
\end{table}

\paragraph{Sampled control on the same checkpoints.}
Table~\ref{tab:8b} reports greedy evaluation. As an additional control, we evaluate the same checkpoints at temperature 0.5 on the 500 AC48 puzzles, comparing eight decodes of the original presentation (pseudo-orbit) with one decode of each of the eight presentations (real orbit).

\begin{table}[htbp]
\centering\small
\caption{\textbf{8B sampled decoding control.} ALL/ANY and gaps (ANY$-$ALL) in percent. Gap increase is the real-orbit gap minus the pseudo-orbit gap.}
\label{tab:8b-control}
\begin{tabular}{lcccc}
\toprule
Recipe & Same input & Real orbit & Real gap & Gap increase\\
\midrule
Canonical & 87.6/87.6 & 48.2/99.8 & 51.6 & 51.6\\
PR & 89.4/89.4 & 49.8/100.0 & 50.2 & 50.2\\
RSM & 94.4/94.6 & 74.6/99.8 & 25.2 & 25.0\\
RSM+PR & 98.2/98.2 & 87.2/100.0 & 12.8 & 12.8\\
\bottomrule
\end{tabular}
\end{table}

In Table~\ref{tab:8b-control}, the observed same-input correctness gaps are 0.0--0.2 points, while real-orbit gaps are 12.8--51.6. PR reduces the real-orbit gap by 1.4 points over canonical SFT and by 12.4 points over RSM. RSM+PR has the highest ALL in both the original greedy evaluation and this sampled evaluation. The density-matched 8B comparison below isolates the contribution of reasoning-state structure.

\paragraph{Density-matched 8B comparison.}
To isolate reasoning-state structure at 8B, we construct one matched random state for each cached RSM state. It keeps the puzzle, the number of hidden cells, and the number of up-weighted targets, but chooses the hidden cells uniformly among the original blanks and the up-weighted targets uniformly among the hidden cells. Arm A trains on these states in the canonical orientation, and arm B adds the same eight-view expansion as RSM+PR. All other settings follow the corresponding RSM branches, including the LoRA configuration, warm start, learning-rate schedule, and seed. Arms A, B, and D are evaluated at 3,500 steps; C is the validation-selected RSM checkpoint of Table~\ref{tab:8b}. Evaluation follows Table~\ref{tab:8b}: greedy decoding on 500 Sudoku-AC48 puzzles with eight views.

\begin{table}[htbp]
\centering\small
\caption{\textbf{Density-matched 8B comparison} (LLaDA-8B-Instruct, LoRA, Sudoku-AC48, percentages).}
\label{tab:8b-matched}
\begin{tabular}{cl ccc}
\toprule
Arm & States / presentation & Acc. & \textsc{Any} & \textsc{All}\\
\midrule
A & Matched random / canonical & 86.8 & 100.0 & 47.0\\
B & Matched random / PR & 88.5 & 99.8 & 52.6\\
C & RSM / canonical & 94.0 & 100.0 & 74.8\\
D & RSM / PR & \textbf{96.6} & 100.0 & \textbf{86.1}\\
\bottomrule
\end{tabular}
\end{table}

In all-view correctness (Table~\ref{tab:8b-matched}), RSM+PR exceeds matched random+PR by 33.5 points, and RSM exceeds matched random states by 27.8 points without PR. The interaction $(D-C)-(B-A)$ is $+5.7$ points, in the same direction as at 50M and 100M.

\subsection{Positional-index pilot}
\label{app:positional-pilot}

Theorem~\ref{thm:rigidity} characterizes which positional indexings certify $D_4$-equivariance. In a small pilot, we replace the row-major RoPE index of a small MDLM with indexings tied to $D_4$ orbits: the three row-major offset classes of the theorem (tied 1D), or the 45 unordered displacement classes $\{|\Delta r|, |\Delta c|\}$ of a $9\times 9$ grid (tied 2D). All other components, data, and optimization are held fixed.

\begin{table}[htbp]
\centering
\caption{\small \textbf{Positional-index pilot} (200 states, $K=8$; percentages except sECR, a fraction).}
\label{tab:positional-pilot}
\vskip 0.1in
\begin{tabular}{l l cccc}
\toprule
Scale & Positional index & MeanSucc & $\widehat C_{\max}$ & $\widehat R_{\min}$ & sECR \\
\midrule
5M & Row-major RoPE & 8.7 & 23.6 & 0.6 & 0.026 \\
5M & Tied 1D (3 classes) & 22.2 & 42.1 & 6.8 & 0.162 \\
5M & Tied 2D (45 classes) & \textbf{25.6} & \textbf{46.8} & \textbf{8.9} & \textbf{0.191} \\
\midrule
10M & Row-major RoPE & 19.0 & 38.4 & 5.1 & 0.132 \\
10M & Tied 1D (3 classes) & 24.9 & 46.7 & 8.2 & 0.175 \\
10M & Tied 2D (45 classes) & \textbf{28.9} & \textbf{50.1} & \textbf{11.1} & \textbf{0.222} \\
\bottomrule
\end{tabular}
\end{table}

Symmetry-tied indexing raises worst-view success and sECR at both scales, and enlarging the certified vocabulary from 3 to 45 classes helps further. The tied models also solve more puzzles on average, so this pilot does not separate stability from competence; it indicates that positional mechanisms respecting the task symmetry are a promising complement to RSM+PR.

\section{Robustness of the RSM--PR Interaction to Matched Controls}
\label{app:matched}\label{sec:matched}\label{app:controls}

The recipe study in Section~\ref{sec:experiments} compares complete training recipes, which differ in their puzzle subsets, masking densities, and target weighting. This appendix holds these factors fixed and asks two questions: whether the RSM--PR interaction persists, and what reasoning-state structure contributes beyond the distribution of partial states it induces.

\subsection{Design}
\label{app:matched-design}

For puzzle $e$, let $T_e$ be its number of cached nonterminal reasoning states and $N_e$ its number of initially hidden cells. All arms sample uniformly from the pooled state cache, so that
\[
\nu(e)=\frac{T_e}{\sum_jT_j},\qquad \pi_e(t)=\frac1{T_e}.
\]
Arms A and B hide a uniformly random subset of original blanks with the same cardinality as the sampled reasoning state; original clues remain visible. Arms C and D use the reasoning state itself. Arm E uses the random subset of B but reweights each target by the cumulative coefficient $\omega_i$ of Eq.~(\ref{eq:omega}):
\[
\ell_E=\frac1{|H|}\sum_{i\in H}N_e\,\omega_i\,\mathrm{CE}_i .
\]
Because $\Pr(i\in H\mid |H|)=|H|/N_e$, the expected coefficient of each target equals $\omega_i$; E therefore matches marginal target weights while randomizing which cells are visible together. A and C use the canonical orientation, and B, D, and E draw one view uniformly from $D_4$ per instance, transporting inputs, targets, and coefficients together.

The original A--E studies start from a pretrained checkpoint and train three fine-tuning seeds per arm for 9,000 updates at batch size 512, using AdamW with learning rate $10^{-5}$, 150 warmup steps, a final learning rate of $10^{-6}$, weight decay 0.01, and gradient clipping at 1.0. The cache contains 67,551 states from 9,995 of 10,000 puzzles (five empty trajectories are excluded). All arms use dense CE over every still-hidden cell at diffusion time 1.0, and we evaluate the final checkpoint. Evaluation uses the same 500 AC24 inputs for every run and scores the full completed board; the decoder uses 128 steps, temperature 1.0, and $K=16$ samples per view. Intervals are paired bootstrap intervals over test puzzles, conditional on the trained runs.

\subsection{Original three-seed A--E results}
\label{app:matched-results}

\begin{table}[htbp]
\centering\small
\caption{\textbf{Matched Sudoku study.} Three-seed averages, in percent. Mean: MeanSucc$^{16}$.}
\label{tab:matched-50m}
\resizebox{\linewidth}{!}{
\begin{tabular}{cl rrr rrr}
\toprule
&&\multicolumn{3}{c}{50M}&\multicolumn{3}{c}{100M}\\
\cmidrule(lr){3-5}\cmidrule(lr){6-8}
Arm & States / presentation & Mean & $\widehat R_{\min}$ & sECR & Mean & $\widehat R_{\min}$ & sECR\\
\midrule
A & Random / canonical & 44.31 & 26.60 & 42.98 & 40.00 & 22.88 & 39.52\\
B & Random / PR & 42.86 & 27.01 & 45.74 & 39.17 & 23.63 & 42.73\\
C & RSM / canonical & 43.63 & 24.47 & 38.61 & 38.05 & 19.48 & 33.39\\
D & RSM / PR & 43.82 & \textbf{27.67} & 45.67 & 39.49 & \textbf{23.84} & 42.76\\
E & Weight-matched random / PR & 42.97 & 27.07 & 45.81 & 39.14 & 23.69 & 43.02\\
\bottomrule
\end{tabular}}
\end{table}

\begin{table}[htbp]
\centering\small
\caption{\textbf{Paired contrasts in the matched study} (percentage points, 95\% intervals). The interaction is $(D-C)-(B-A)$. $^\dagger$Point estimate from Table~\ref{tab:matched-50m}.}
\label{tab:matched-contrasts}
\resizebox{\linewidth}{!}{
\begin{tabular}{llcc}
\toprule
Contrast & Measure & 50M & 100M\\
\midrule
C$-$A & sECR & $-4.36\ [-5.75,-3.07]$ & $-6.13\ [-7.50,-4.78]$\\
C$-$A & $\widehat R_{\min}$ & $-2.13^\dagger$ & $-3.40\ [-4.19,-2.62]$\\
Interaction & sECR & $+4.30\ [+2.88,+5.76]$ & $+6.16\ [+4.63,+7.73]$\\
Interaction & $\widehat R_{\min}$ & $+2.80\ [+1.96,+3.63]$ & $+3.60\ [+2.77,+4.44]$\\
Interaction & Mean & $+1.64\ [+1.14,+2.12]$ & $+2.27\ [+1.72,+2.84]$\\
D$-$B & Mean & $+0.96\ [+0.62,+1.33]$ & $+0.32\ [-0.03,+0.68]$\\
D$-$B & $\widehat R_{\min}$ & $+0.66\ [+0.10,+1.22]$ & $+0.20\ [-0.34,+0.74]$\\
D$-$B & sECR & $-0.07\ [-1.01,+0.83]$ & $+0.04\ [-1.01,+1.06]$\\
D$-$E & Mean & $+0.85\ [+0.52,+1.20]$ & $+0.35\ [+0.00,+0.71]$\\
D$-$E & $\widehat R_{\min}$ & $+0.60\ [+0.04,+1.15]$ & $+0.15\ [-0.39,+0.69]$\\
D$-$E & sECR & $-0.14\ [-1.06,+0.75]$ & $-0.26\ [-1.29,+0.79]$\\
\bottomrule
\end{tabular}}
\end{table}

\paragraph{View exposure is most valuable for reasoning-state supervision.}
Without PR, reasoning states lower sECR by 4.36 points at 50M and 6.13 points at 100M relative to random states with the same puzzle frequencies and hidden counts, and lower estimated worst-view success by 2.13 and 3.40 points. This reproduces, under matched supervision, the pattern of Section~\ref{sec:experiments}: reasoning-state supervision on its own amplifies view dependence.

\paragraph{The interaction persists at both scales.}
PR raises sECR by 7.06 points under RSM and 2.76 points under random states at 50M (9.37 and 3.21 at 100M), giving interactions of $+4.30$ and $+6.16$ with intervals well above zero. The interaction is equally clear in worst-view success ($+2.80$, $+3.60$) and mean success ($+1.64$, $+2.27$). The benefit of view exposure thus depends on the supervision it accompanies, and it is largest when paired with reasoning states.

\paragraph{What reasoning-state structure adds.}
The original augmented endpoints are close in sECR, with intervals that do not establish an additional normalized-retention gain. The 50M mean-success and estimated-worst-view comparisons favor D over B/E; the 100M D--B mean comparison is unresolved. These finite-$K$ endpoints neither imply equivalence nor replace the factorial question. At 8B, the corresponding density-matched comparison favors reasoning states by a large margin in all-view correctness (Appendix~\ref{app:8b}). The expanded design and population-dispersion analysis follow.

\subsection{Expanded factorial and endpoint contrasts}
\label{app:matched-extension}
Arm F is the canonical counterpart of E, with exactly the same random-state sampler and weights. Thus F/E are target-weight-matched random states without/with PR, whereas C/D are reasoning states without/with PR. All augmented arms use the same online uniform-view procedure. The extension has five fine-tuning seeds per arm at 50M and three at 100M. We distinguish this from the original three-seed A--E endpoints in Tables~\ref{tab:matched-50m}--\ref{tab:matched-contrasts}; those original outputs are retained, not relabeled as five-seed results.

\begin{table}[htbp]
\centering\small
\caption{\textbf{Expanded matched factorial.} Percentage-point contrasts; intervals resample puzzles conditional on the listed training runs unless explicitly marked crossed. $I=(D-C)-(B-A)$ and $I^{\rm wt}=(D-C)-(E-F)$.}
\label{tab:matched-expanded}
\begin{tabular}{llcc}
\toprule
Contrast & Metric & 50M (5 seeds) & 100M (3 seeds)\\
\midrule
$I$ & Mean & $+1.57\ [1.09,2.07]$ & $+2.27\ [1.72,2.84]$\\
$I$ & sECR & $+4.03\ [2.65,5.44]$ & $+6.17\ [4.57,7.73]$\\
$I^{\rm wt}$ & Mean & $+1.51\ [1.04,1.98]$ & $+2.45\ [1.89,3.01]$\\
$I^{\rm wt}$ & sECR & $+4.06\ [2.61,5.56]$ & $+6.32\ [4.70,7.92]$\\
$E-F$ & Mean & $-1.29\ [-1.65,-0.94]$ & $-1.00\ [-1.37,-0.64]$\\
$D-C$ & Mean & $+0.22\ [-0.31,0.76]$ & $+1.44\ [0.84,2.04]$\\
$D-F$ & Mean & $-0.46\ [-0.86,-0.07]$ & $-0.65\ [-1.10,-0.21]$\\
\bottomrule
\end{tabular}
\end{table}
The 50M density-matched mean endpoint D--B in the five-seed extension is $+0.94\ [0.57,1.29]$ points. The original three-seed density- and weight-matched mean gains are $+0.96\ [0.62,1.33]$ and $+0.85\ [0.52,1.20]$. Their estimated worst-view gains, $+0.66\ [0.10,1.22]$ and $+0.60\ [0.04,1.15]$, are secondary finite-$K$ diagnostics. The 100M D--B mean endpoint is $+0.32\ [-0.03,0.68]$, unresolved. D does not dominate every unaugmented control: the D--F mean contrasts are negative in Table~\ref{tab:matched-expanded}. An interaction and a positive augmented-endpoint contrast are distinct claims.

For the unaugmented state comparison, original C--A sECR differences are $-4.36$ at 50M and $-6.13$ at 100M. With the added F control, 100M C--F is $-6.57\ [-7.95,-5.18]$. The latter must not be labeled as C--A. Point-estimate differences computed from rounded arm means can differ from reported unrounded contrasts.

\subsection{Finite-sample measurements and training uncertainty}
\label{app:measurement-results}
The A--D success-dispersion interaction $J$ from Eq.~\ref{eq:unbiased-v} is $+0.00289\ [0.00185,0.00394]$ at 50M and $+0.00440\ [0.00338,0.00548]$ at 100M. These are not automatically the fully target-weight-matched C/D/F/E contrasts. Under the stronger mutually independent Bernoulli null, arm-wise conditional tests reject equal view success at Monte Carlo resolution $p=0.001$. This is a joint equal-view null within an arm, not a direct test of $J>0$. A separate pooled-success plug-in extrema reference leaves sECR interactions of $+2.96$ and $+4.45$ points. These calibrations are not literal decompositions of an arbitrary finite-$K$ effect into signal and artifact. No empirical large-$K$ convergence is asserted from these results.

\paragraph{Token-marginal calibration at 50M.}
For A--D and training seeds 0--4, each puzzle contributes 128 whole canonicalized grids. We repartition individual grids into eight pseudo-views of sixteen, 400 times with analysis seed 20261108. The generator was initialized once and not reset across views; the audited path does not share a replicate-noise draw across views. Under the exchangeable-decoding null, the permutation reference of Appendix~\ref{app:token-null} applies. Recomputed observed means agree with the saved statistic within $10^{-6}$.
\begin{table}[htbp]
\centering\small
\caption{\textbf{50M token-marginal calibration, five seeds.} Units are nats. Excess is observed minus the whole-output repartition reference. Arm intervals resample 500 paired puzzles while retaining the seed blocks.}
\label{tab:token-calibration}
\begin{tabular}{lrrr}
\toprule
Arm & Observed & Null mean & Excess [95\% interval]\\
\midrule
A random/canonical & 0.07595 & 0.06942 & $0.00653\ [0.00575,0.00735]$\\
B random/PR & 0.07271 & 0.06999 & $0.00272\ [0.00210,0.00334]$\\
C RSM/canonical & 0.08003 & 0.06769 & $0.01235\ [0.01113,0.01359]$\\
D RSM/PR & 0.07203 & 0.06784 & $0.00419\ [0.00344,0.00494]$\\
\bottomrule
\end{tabular}
\end{table}
The raw interaction $(D-C)-(B-A)$ is $-0.00476\ [-0.00597,-0.00355]$. The excess interaction is $-0.00434\ [-0.00543,-0.00325]$, or $[-0.00547,-0.00322]$ with crossed seed-by-puzzle resampling. All five seed-specific excess interactions are negative. PR reduces excess by 0.00816 under RSM and 0.00382 under random states. The Monte Carlo SE of the seed-averaged interaction is $1.3\times10^{-5}$. Lower leakage is better: this is a greater reduction under RSM, not a claim that D's excess endpoint (0.00419) is below B's (0.00272). All values derive from unrounded arrays; the calibrated statistic is not an unbiased estimator of population KL.

\paragraph{Independent fine-tuning repetitions.}
The five 50M density-matched mean interactions are 1.70, 1.65, 1.57, 1.46, and 1.50 points (between-seed SD 0.098). The puzzle-only interval is $[1.09,2.07]$ and the crossed interval $[1.08,2.07]$. The three 100M weight-matched mean interactions are 2.636, 2.455, and 2.248. These measure fine-tuning variation at the specified pretrained initialization; checkpoints of one pretraining trajectory are not additional independent seeds. The full recipe repetition is separately reported in Appendix~\ref{app:recipe-seeds}.

\subsection{Predefined Sudoku difficulty strata}
\label{app:difficulty}
Strata are defined from test inputs before examining model contrasts: singles-only completion (130 puzzles), fallback with wave count below/above its median (57/41), and oracle-unfinished (272). In the unfinished stratum, D--B mean success is $+1.22\ [0.74,1.72]$ points and D--E is $+1.15\ [0.67,1.64]$. The separate difficulty-slope contrast, $+0.20\ [-0.07,0.48]$, is unresolved; we do not infer monotone growth of benefit with difficulty. This is a Sudoku analysis, not the maze slope.

\subsection{Presentation-refresh interventions and alternatives}
\label{app:mechanism-results}
All samplers are measured at the same pretrained 50M checkpoint using the last two blocks and head. The raw projected $W_R-W_0$ is $1.78\times10^{-4}\ [1.51\times10^{-4},2.07\times10^{-4}]$, approximately an eleven-fold trace ratio. Common-denominator normalization and measurements at the final canonical checkpoints retain the positive difference. Frozen nested-random and independent-random banks account for about 3\% and 2\% of this R-versus-online gap, respectively. These are projected learning-signal measurements, not full-model trace estimates.

\begin{table}[htbp]
\centering\small
\caption{\textbf{Refresh intervention.} Three fine-tuning seeds per sampler and schedule. The schedule labels follow the experiment's fixed/intermediate/fresh design; they are not substituted for the exact finite-$n$ coefficient $\alpha$. Values are sECR interaction points.}
\label{tab:refresh}
\begin{tabular}{lrrrr}
\toprule
Schedule & Fixed & Intermediate 1 & Intermediate 2 & Fresh\\
\midrule
Interaction & 0.06 & 2.44 & 3.13 & 4.30\\
\bottomrule
\end{tabular}
\end{table}
The intermediate difference is $+0.69\ [-0.51,1.92]$, unresolved. The reported rank association in the common-response diagnostic is Kendall $\tau=0.87\ [0.60,1.00]$ for sECR and $0.60\ [0.33,0.87]$ for success dispersion; it does not establish a universal quantitative mapping from $W$ to performance.

A frozen nested-random state bank uses no reasoning solver. Its canonical sECR cost relative to online random states is $-1.12\ [-2.09,-0.19]$, and its state--PR interaction is $+1.74\ [0.66,2.90]$. The independent-random frozen bank is measured at the gradient level but is not a completed trained-model arm. Freezing and nesting can therefore contribute; reasoning semantics alone do not guarantee presentation stability.

The experimental alternatives are retained. An axis-specific subgroup prediction was not confirmed: the two subgroup recovery summaries were 0.73 and 0.71 and their interaction contrast includes zero. A fixed random view alone changes RSM sECR by $+1.17\ [-0.24,2.60]$, unresolved. The nested control's behavioral cost is not proportional to its roughly 3\% share of the persistence gap, so a quantitative proportionality hypothesis is not confirmed. Finally, the hypothesis that unaugmented RSM has a hidden advantage in its training presentation is not supported: at 100M, identity-view C--A mean success is $-2.83\ [-4.02,-1.67]$ and the nonidentity average is $-1.83$. These delimit the interpretation rather than being discarded from the mechanism account.

\section{Experimental details}
\label{app:exp-details}

This appendix gives the details needed to reproduce the experiments in Section~\ref{sec:experiments}.  The main paper states the logic of each experiment.  Here we state how the data, parser, sampling controls, and uncertainty estimates are implemented.

\subsection{Anchored-completion generation}
\label{app:exp-generation}

We generate anchored-completion states only after the base Sudoku puzzle split.  For each base puzzle, we choose anchors $A$, target cells $U$, and unresolved cells $R$.  We keep the state only if the values of $U$ are forced by $A$:
\begin{equation}
\{s_U:s\in\mathcal S(A)\}=\{s_U^\star\}.
\end{equation}
The verification uses a Sudoku solver that enumerates completions until either two different target assignments are found or the target assignment is certified unique.  The full grid need not be unique.  Only the target assignment must be unique.

After a state is retained, we apply all eight elements of $D_4$.  Because rotations and reflections preserve Sudoku row, column, and box constraints, every transformed state has the corresponding transformed target.  During scoring, every prediction is mapped back to canonical coordinates before comparison.

\subsection{Parsing and scoring}
\label{app:exp-parsing}

The model is asked to output the values of the target cells.  A prediction is valid only if it can be parsed into exactly one digit for each requested target cell.  If the output is malformed, incomplete, contains an invalid digit, or assigns two different values to the same target cell, it is mapped to the invalid symbol $\bot$ and scored as incorrect.

Invalid outputs contribute failures to every correctness summary. The group action fixes the invalid symbol: $g\bot=\bot$.

\subsection{Stochastic sampling}
\label{app:exp-sampling}

For stochastic evaluation, we draw $K=16$ samples per view unless otherwise stated.  For each base state $x$ and view $g$, the empirical success probability is
\begin{equation}
\widehat p_{g,K}(x)=\frac1K\sum_{k=1}^{K}\mathbf 1[g^{-1}Y_{g,k}=y^\star(x)].
\end{equation}
We then compute
\begin{equation}
\widehat C^K_{\max}(x)=\max_{g\in G}\widehat p_{g,K}(x),
\qquad
\widehat R^K_{\min}(x)=\min_{g\in G}\widehat p_{g,K}(x).
\end{equation}
The extrema are computed within each base state and then averaged over base states.  We do not pool transformed views as independent examples.

\subsection{Canonical pseudo-orbit control}
\label{app:exp-pseudo-orbit}

The pseudo-orbit control measures the extrema gap produced by repeated sampling from the same canonical input.  For each base state, we draw $|G|K$ samples from the canonical layout, randomly split them into $|G|$ groups of size $K$, and compute the same extrema used for the real orbit:
\begin{equation}
\widehat C^{K,\mathrm{pseudo}}_{\max}(x),
\qquad
\widehat R^{K,\mathrm{pseudo}}_{\min}(x).
\end{equation}
The pseudo-orbit has no true geometric view shift. The resulting contrast compares cross-view behavior with this same-input reference, but is not an unbiased correction under arbitrary unequal view success rates. The additional large-model $m$-call ANY/ALL experiment in Appendix~\ref{app:decoding-controls} is a distinct protocol from this $mK$-sample extrema control.

\subsection{Token view-leakage diagnostic}
\label{app:exp-diagnostics}

For each target cell $i$ and view $g$, we estimate a token distribution $\widehat p_{g,i}$ from the $K$ canonicalized samples.  We add a small symmetric smoothing constant $\epsilon$ to every digit count before normalization.  The reported token view-leakage proxy is
\begin{equation}
\widehat L_{\mathrm{tok}}(x)
=
\frac1{|U|}\sum_{i\in U}\frac1{|G|}\sum_{g\in G}
D_{\mathrm{KL}}(\widehat p_{g,i}\Vert \widehat{\overline p}_i),
\qquad
\widehat{\overline p}_i=\frac1{|G|}\sum_{g\in G}\widehat p_{g,i}.
\end{equation}
This token-marginal diagnostic can be positive under finite sampling even when the generating laws agree. Agreement of token marginals does not establish equality of full output laws. The whole-output repartition reference in Appendix~\ref{app:token-null} uses the same canonicalization, target cells, smoothing, and KL implementation. It moves individual complete outputs, not individual tokens or entire view bins. For the five-seed 50M experiment there are 400 repartitions per puzzle; all reported intervals use base-puzzle resampling, with crossed intervals additionally resampling the seed blocks.

\subsection{Bootstrap intervals and paired comparisons}
\label{app:exp-bootstrap}

The original Sudoku intervals use 10,000 paired bootstrap resamples of base puzzles, retaining all views and repeated decodes together. They condition on the trained runs. The new recipe analysis uses analysis seed 20261108 and percentile 2.5--97.5\% intervals. Its crossed bootstrap samples three seed-block indices once and uses them for all four arms; it likewise samples puzzle indices once across arms. For each resampled seed, it recomputes the ratio of puzzle-averaged extrema, then averages seed-specific ratios. Interactions are computed within each bootstrap replicate, never by subtracting interval endpoints.

The matched extension separately reports five fine-tuning seeds at 50M and three at 100M. Original three-seed results are kept identifiable. The matched-maze interval resamples seed blocks and mazes. Independence here concerns distinct fine-tuning runs conditional on the stated pretrained initialization; no count of checkpoints is used as a count of pretraining seeds. Few seed blocks limit seed-level precision, so individual contrasts and between-seed SD accompany intervals. These percentile intervals are not claimed to have exact finite-sample coverage.
\subsection{Pretraining}
\label{app:exp-pretrain}

Controlled MDLMs use DiT backbones with SUBS parameterization and sequence length 81 (one token per cell). Pretraining runs for 500K steps (5M, 10M, 50M) or 700K steps (100M) with global batch size 512 (256 for 100M), and we select the best checkpoint on validation. Sampling-based evaluation fixes the random seed to 0.

\subsection{Post-training}
\label{app:exp-posttrain}

The initial 50M recipes start from one pretrained checkpoint and train for 9,000 updates with batch size 512 and learning rate $10^{-5}$. RSM uses dense completion over every still-hidden target, not just the next solver frontier; SFT baselines use their random-corruption pipeline. The original PR preset \texttt{d4\_shuffle10k\_indmask}/\texttt{full\_orbit\_sft\_10k} expands 10,000 puzzles to all eight D4 views, globally shuffles the 80,000 examples, and applies independent corruption after transformation. PR grouped keeps all eight views of a puzzle together and averages the losses. RSM+PR uses state-orbit expansion. The matched factorial instead samples a uniform D4 view online in each augmented arm, with the same procedure across those arms.

\begin{table}[htbp]
\centering\small
\caption{\textbf{One state generator, two retention rules.} Five of 10,000 puzzles emit no retained states. Complete-only retention is not synonymous with emitting any trajectory.}
\label{tab:cache-retention}
\begin{tabular}{llrr}
\toprule
Study & Retention rule & Puzzles & States\\
\midrule
Initial recipe & Trajectories reaching completion & 4,633 & 41,135\\
Additional retained states & Valid unfinished trajectories & 5,362 & 26,416\\
Matched study & Both subsets & 9,995 & 67,551\\
\bottomrule
\end{tabular}
\end{table}
The added recipe seeds use the complete-only cache, not the 67,551-state matched cache. RSM settings use target masks, dense loss, frontier weight 1, no online random mask, and state-orbit expansion only for RSM+PR. The additional runs use batch 512, learning rate $10^{-5}$, warmup 150, and 9,000 updates. Final checkpoints are evaluated with batch 64, generator seed 0 initialized once and not reset per view, exact match, 128 steps, and temperature 1. Different evaluation batch sizes can change sampled outputs; the new common-harness values are not asserted to reproduce the original samples exactly.

LLaDA-8B-Instruct uses LoRA, 8,000 optimizer steps, and validation selection. Every 500 steps, 200 validation puzzles are decoded greedily, and the checkpoint with the highest exact-grid-match accuracy is selected. The AC48 comparison uses one greedy prediction per view, and the temperature-0.5 pseudo-orbit control uses the same checkpoints. The density-matched 8B arms are described in Appendix~\ref{app:8b}.LLaDA-8B-Instruct uses LoRA, 8,000 optimizer steps, and validation selection. Every 500 steps, 200 validation puzzles are decoded greedily, and the checkpoint with the highest exact-grid-match accuracy is selected. The AC48 comparison uses one greedy prediction per view, and the temperature-0.5 pseudo-orbit control uses the same checkpoints. The density-matched 8B arms are described in Appendix~\ref{app:8b}.

\subsection{Paired code and mathematics evaluations}
\label{app:exp-paired}

HumanEval-Infilling pairs are generated by applying one semantics-preserving edit to the surrounding context: identifier renaming, adjacent-statement swap, statement relocation, operand exchange with operator dualization, or inert-line insertion. A pair is admitted only if the transformed program passes the original test harness and matches the original's outputs. MATH-500 transformations (variable renaming, premise permutation, identity-operation insertion) are admitted only if they preserve the canonical answer. Predictions are aligned by the transformation before scoring, and \textsc{Any}/\textsc{All} are computed over each original--transformed pair.

\subsection{Decoding audits and aggregation}
\label{app:decoding-audit}
The original large-model and 8B tables use greedy decoding: temperature 0 with argmax token selection. LLaDA-family models use low-confidence unmasking with top-$k$ confidence positions, and Dream uses entropy-based remasking.

The cross-task pseudo-orbit controls use temperature 0.5 for Sudoku and code and 0.2 for MATH. All original and transformed texts are scored with the same task checker within a comparison. Repeats of one original input and changes of presentation are kept distinct. The output-level token calibration uses a generator initialized once, with no reset per orbit transform and no shared replicate-noise draw across views.

MATH code/notation changes are pooled only as admitted pairs for the original dLLM tables. The new AR orbit table instead gives each base problem equal weight across all of its available presentations; Appendix~\ref{app:ar-ladder} also reports pair-weighted values explicitly. Parsing failures are not dropped. All new study summaries are kept separate from original outputs when sample selection, prompting, temperature, or evaluation batching differs.

\subsection{Projected persistence measurement and refresh controls}
\label{app:persistence-estimation}
For $W_a$ in Eq.~\ref{eq:persistence-matrix}, sample puzzles proportional to $\nu(e)^2$. At a common checkpoint, obtain two independent sets of states for each puzzle and form separate estimates $\widehat r_e^{(1)}(g)$ and $\widehat r_e^{(2)}(g)$. The split-sample inner product
\[
\widehat c_e=m^{-1}\sum_g\langle\widehat r_e^{(1)}(g),\widehat r_e^{(2)}(g)\rangle
\]
is unbiased for $\E_g\|r_{a,e}(g)\|^2$. Multiply its average by $\sum_e\nu(e)^2$ to estimate $W_a$. Draw states with replacement from fixed banks; do not square a single noisy gradient estimate. Negative split-sample estimates are retained. All gradients are projected to the last two blocks plus head. Raw values are primary; any normalization uses a denominator shared across samplers.

The behavioral refresh suite trains reasoning and random-state models at fixed, two intermediate, and fresh presentation schedules with three seeds. The exact theorem refers to block lengths through $\alpha$, and schedule labels alone are not substituted into the theorem. The nested-random bank preserves marginal hidden counts but selects nested prefixes from a random reveal order. Results and non-confirmed alternatives are in Appendix~\ref{app:mechanism-results}.

\section{Reasoning States and the Supervision Objective}
\label{app:general_lts}

RSM selects conditional inputs rather than prescribing an output order. For a fixed problem with $N$ initially hidden targets, unrestricted masking admits $2^N-1$ nonterminal visibility patterns. A single monotone trajectory that reveals at least one new target per transition contains at most $N$ nonterminal states. This counts selected inputs; it is not a sample-complexity statement. Different states can supervise the same target under different visible information.

If a revealed assignment is already entailed by the visible facts, adding it need not shrink the set of logically feasible solutions. It makes an intermediate conclusion explicit. Dense RSM supervision then predicts every unresolved target from that input, rather than only the next frontier. Analogous task encodings can use verified intermediate facts or program-execution values; the training experiments here instantiate RSM on grids.

\subsection{The orbit-averaged completion objective}
\label{app:opinion}
For one state--target pair, let $q_g$ be positive canonicalized predictions on a common finite label alphabet, and let
\[
\widetilde r(a)=\prod_gq_g(a)^{1/m},\quad Z=\sum_a\widetilde r(a),\quad r(a)=\widetilde r(a)/Z.
\]
The logarithmic-opinion-pool identity is
\begin{equation}
\frac1m\sum_g-\log q_g(y)
=-\log r(y)+\frac1m\sum_g D_{\mathrm{KL}}(r\Vert q_g).
\label{eq:ambiguity}
\end{equation}
Indeed, $\log r=\frac1m\sum_g\log q_g-\log Z$, so the average reverse KL equals $-\log Z$. RSM selects the state--target distribution over which this identity is averaged. Minimizing the sum does not require both summands to decrease on every update, and the identity applies to random-state supervision as well. The geometric consensus is an analytical quantity; inference does not ensemble over views.

\subsection{Presentation persistence and random objectives}
\label{app:persistence-theory}
The following results formalize the sampling intervention in Section~\ref{sec:persistence}. They assume a fixed finite puzzle pool, puzzle probabilities $\nu(e)$, independent state draws conditional on puzzles, presentation schedules independent of states, and finite second moments. Differentiation and expectation are interchanged under the usual integrability condition. These are assumptions on the sampling protocol and loss, not a convergence assumption on a transformer.

For sampler $a$, define
\begin{align*}
v_{e,s}(g;\theta)&=\nabla_\theta\ell_\theta(g\cdot z_{e,s}),\\
m_{a,e}(g;\theta)&=\E_{s\sim\pi_{a,e}}v_{e,s}(g;\theta),\\
r_{a,e}(g;\theta)&=m_{a,e}(g;\theta)-\E_hm_{a,e}(h;\theta).
\end{align*}
Here $g$ transports inputs, target indices, labels, and any target weights. The matrices $C_a$ and traces $W_a$ are defined in Eq.~\ref{eq:persistence-matrix}. The squared puzzle weights arise from repeated visits to the same puzzle. For a frozen state bank, $m_{a,e}$ is its weighted bank average.

\begin{theorem}[Fixed random presentations change the objective]
\label{thm:random-objective}
Let $U=(U_e)_e$ assign independent uniform presentations to puzzles, and define
\[
L_{a,U}(\theta)=\sum_e\nu(e)\E_s\ell_\theta(U_e\cdot z_{e,s}),\quad
\bar L_a(\theta)=\E_U L_{a,U}(\theta),\quad
\Delta_{a,U}=L_{a,U}-\bar L_a.
\]
Then $\nabla\Delta_{a,U}(\theta)=\sum_e\nu(e)r_{a,e}(U_e;\theta)$. For each parameter vector fixed independently of $U$,
\[
\E_U\nabla\Delta_{a,U}(\theta)=0,\qquad
\operatorname{Cov}_U(\nabla\Delta_{a,U}(\theta))=C_a(\theta).
\]
Along SGD, let $\mathcal F_t$ contain the history before the next sample. With fixed $U$,
\[
\E[\widehat g_t\mid\mathcal F_t,U]=\nabla L_{a,U}(\theta_t);
\]
with fresh independent uniform presentations,
\[
\E[\widehat g_t\mid\mathcal F_t]=\nabla\bar L_a(\theta_t).
\]
\end{theorem}
\begin{proof}
Differentiate the finite weighted sum and subtract its view average to obtain the expression for $\nabla\Delta$. Each $r_{a,e}$ is centered under a uniform view. Independent view assignments make cross-puzzle covariance terms vanish, leaving $\sum_e\nu(e)^2\E_g[r_{a,e}(g)r_{a,e}(g)^\top]$. Conditional on the training history (and on $U$ in the fixed-table case), the next independent puzzle and state draws have laws $\nu$ and $\pi_{a,e}$. Averaging their gradient gives $\nabla L_{a,U}(\theta_t)$. Averaging also over a fresh uniform view gives $\nabla\bar L_a(\theta_t)$.
\end{proof}
The perturbation is a function of $\theta$, not a constant vector. The zero-mean identity at a fixed parameter vector must not be substituted at $\theta_t(U)$ as though the learned parameters were independent of the view table. The conditional-gradient statements above are the claims that hold along the actual trajectory. Nor is an all-identity canonical table an independent uniform table; conclusions for random tables do not erase canonical-frame effects.

\subsection{Exact block-refresh covariance}
\label{app:block-refresh-proof}
For completeness, define draw $j$ in block $b$ as
$G_j=v_{E_j,S_j}(U_{b,E_j};\theta)$, with $E_j\overset{\mathrm{iid}}\sim\nu$ and independent conditional state draws. All $U_{b,e}$ are mutually independent uniform presentations. Let $\widehat G_a=n^{-1}\sum_jG_j$ and $\Sigma_a=\operatorname{Cov}(G_j)$, with $\theta$ fixed independently of the random draws.

\begin{proof}[Proof of Theorem~\ref{thm:refresh-main}]
Each diagonal covariance is $\Sigma_a$. For distinct $i,j$ in the same block, condition on its full view table $U_b$. The draws are conditionally independent, with conditional mean
\[
M_a(U_b)=\sum_e\nu(e)m_{a,e}(U_{b,e};\theta).
\]
The total-covariance identity gives
\[
\operatorname{Cov}(G_i,G_j)=\operatorname{Cov}_{U_b}(M_a(U_b))
=\sum_e\nu(e)^2\E_g[r_{a,e}(g)r_{a,e}(g)^\top]=C_a.
\]
Distinct blocks have independent draws and tables, hence zero cross-covariance. There are $n_b(n_b-1)$ ordered off-diagonal pairs in block $b$. Summing and dividing by $n^2$ gives
\[
\operatorname{Cov}(\widehat G_a)=\frac{\Sigma_a}{n}
+\frac{\sum_bn_b(n_b-1)}{n^2}C_a.
\]
\end{proof}
A single fixed random table gives $\alpha=1-1/n$; fresh presentations give $\alpha=0$. The probability that two randomly selected distinct draw indices reuse a view-table draw is
$q_{\rm pair}=\sum_bn_b(n_b-1)/(n(n-1))$. This is not the probability of equal view labels, which also includes accidental agreement and is $q_{\rm pair}+(1-q_{\rm pair})/m$.

\begin{corollary}[State-dependent refresh effect in the learning signal]
\label{cor:gradient-interaction}
Let $\mu_a=\E\widehat G_a$. The gradient mean-squared-error reduction from block refresh coefficient $\alpha$ to fresh PR is $\alpha W_a$. Its difference between samplers R and 0 is $\alpha(W_R-W_0)$.
\end{corollary}
\begin{proof}
Both schedules have the same mean $\mu_a$ and single-draw covariance $\Sigma_a$. Since $\E\|\widehat G_a-\mu_a\|^2=\operatorname{tr}\operatorname{Cov}(\widehat G_a)$, subtracting the fresh covariance gives $\alpha\operatorname{tr}C_a$. Subtract the two samplers' reductions.
\end{proof}
This is an exact statement about gradient MSE at fixed parameters. Neither its sign nor size is a theorem about final sECR, mean accuracy, or augmented-endpoint dominance. The experiment measures those quantities independently.

\subsection{Freezing a state bank can itself increase persistence}
\label{app:frozen-bank}
Let a random finite bank $\mathcal K$ be constructed so that its expected per-puzzle sampling measure equals that of online sampler 0. All expectations below use parameters fixed independently of bank construction.
\begin{theorem}[Frozen-bank persistence]
\label{thm:frozen-bank}
If $\E_{\mathcal K}r_{\mathcal K,e}(g)=r_{0,e}(g)$, then
\[
\E_{\mathcal K}W_{\mathcal K}-W_0
=\sum_e\nu(e)^2\E_{\mathcal K,g}
 \|r_{\mathcal K,e}(g)-r_{0,e}(g)\|^2\ge0.
\]
\end{theorem}
\begin{proof}
Write $r_{\mathcal K,e}=r_{0,e}+\varepsilon_{\mathcal K,e}$ with $\E_{\mathcal K}\varepsilon_{\mathcal K,e}=0$. Expand the squared norm, average over the bank and view, and cancel the mean-zero cross term. Multiply by $\nu(e)^2$ and sum.
\end{proof}
Independent random subsets stored once, or nested prefixes of a random cell order with the same marginal hidden counts, meet the measure-matching condition. A reasoning-state bank need not, so the theorem does not establish $W_R\ge W_0$. Frozen controls separate caching and nesting from reasoning order. The measured nested control supplies a behavioral test, and the untrained independent bank supplies a learning-signal diagnostic; they are not conflated.

\subsection{Scope of the theoretical and behavioral connection}
\label{app:persistence-scope}
The exact results identify the protocol-dependent learning signal. A monotone behavioral response to persistence and a common response curve across state samplers are additional empirical hypotheses. The refresh experiment tests them but does not turn the covariance identity into a final-performance guarantee. The intermediate refresh contrast is unresolved and a quantitative proportionality prediction for nested states is not confirmed (Appendix~\ref{app:mechanism-results}). Gradient measurements use the declared last-two-blocks-plus-head projection, not the full parameter vector. This distinction is retained in the results and interpretation.

\end{document}